%% file: main.tex
\documentclass[english]{article}
\usepackage[T1]{fontenc}
\usepackage[latin9]{inputenc}
\usepackage[final]{neurips_2022}
\usepackage{array}
\usepackage{float}
\usepackage{textcomp}
\usepackage{url}
\usepackage{pdfpages}
\usepackage{multirow}
\usepackage{amsmath}
\usepackage{amsthm}
\usepackage{amssymb}
\usepackage{graphicx}
\usepackage{wasysym}
\PassOptionsToPackage{normalem}{ulem}
\usepackage{ulem}

\makeatletter

\providecommand{\tabularnewline}{\\}

\floatstyle{ruled}
\newfloat{algorithm}{tbp}{loa}
\providecommand{\algorithmname}{Algorithm}
\floatname{algorithm}{\protect\algorithmname}

\usepackage{algorithm}
\usepackage{algorithmic}
\usepackage{hyperref}

\makeatother

\usepackage{babel}
\begin{document}
\title{Learning to Constrain Policy Optimization with Virtual Trust Region }
\author{\textbf{Hung Le, Thommen Karimpanal George, Majid Abdolshah, Dung
Nguyen,}\\
\textbf{ Kien Do, Sunil Gupta, Svetha Venkatesh }\\
Applied AI Institute, Deakin University, Geelong, Australia\\
\texttt{thai.le@deakin.edu.au}}
\maketitle
\begin{abstract}
We introduce a constrained optimization method for policy gradient
reinforcement learning, which uses a \emph{virtual trust region} to
regulate each policy update. In addition to using the proximity of
one single old policy as the normal trust region, we propose forming
a second trust region through another \emph{virtual policy} representing
a wide range of past policies. We then enforce the new policy to stay
closer to the virtual policy, which is beneficial if the old policy
performs poorly. More importantly, we propose a mechanism to automatically
build the virtual policy from \emph{a memory of past policies}, providing
a new capability for dynamically learning appropriate virtual trust
regions during the optimization process. Our proposed method, dubbed
Memory-Constrained Policy Optimization (MCPO), is examined in diverse
environments, including robotic locomotion control, navigation with
sparse rewards and Atari games, consistently demonstrating competitive
performance against recent on-policy constrained policy gradient methods. 
\end{abstract}
\global\long\def\sigmoid{sigmoid}%

\section{Introduction}

\input{intro.tex}

\section{Background: Policy Optimization with Trust Region \label{sec:Background:-Constrained-Policy}}

\input{bg.tex}

\section{Memory-Constrained Policy Optimization}

\input{method.tex}

\section{Experimental results}

\input{exp.tex}

\section{Related work}

\input{related.tex}

\section{Discussion}

\input{discuss.tex}

\section*{Acknowledgments}

This research was partially funded by the Australian Government through
the Australian Research Council (ARC). Prof Venkatesh is the recipient
of an ARC Australian Laureate Fellowship (FL170100006).

\bibliographystyle{plain}
\bibliography{mapi}
\newpage{}

\section*{Appendix}
\renewcommand\thesubsection{\Alph{subsection}}

\input{app.tex}
\end{document}

%% file: intro.tex
Deep reinforcement learning (RL) is the current workhorse in machine
learning. Using neural networks to approximate value and policy functions
enables classical approaches such as Q-learning \cite{watkins1992q}
and policy gradient \cite{sutton2000policy} to achieve promising
results on many challenging problems such as Go, Atari games and robotics
\cite{silver2017mastering,mnih2015human,lillicrap2016continuous,mnih2016asynchronous}.
Compared to Deep Q-learning, deep policy gradient (PG) methods are
often more flexible and applicable to discrete and continuous action
problems. However, these methods tend to suffer from high sample complexity
and training instability since the gradient may not accurately reflect
the policy gain when the policy changes substantially \cite{Kakade2002ApproximatelyOA}.
This is exacerbated for deep policy networks where numerous parameters
need to be optimized, and minor updates in parameter space can lead
to considerable changes in policy space. 

To address this issue, one solution is to regularize each policy update
by restricting the Kullback--Leibler (KL) divergence between the
new policy and the previous one, which can guarantee monotonic policy
improvement \cite{schulman2015trust}. However, jointly optimizing
the approximate advantage function and the KL term does not work in
practice \cite{schulman2015trust}. Therefore, Schulman et al. (2015)
proposed Trust Region Policy Optimization (TRPO) to constrain the
new policy within a KL divergence radius, which requires second-order
gradients. Alternatives such as Proximal Policy Optimization (PPO)
\cite{schulman2017proximal} use a simpler first-order optimization
with adaptive KL or clipped surrogate objective while still maintaining
the reliable performance of TRPO. Recent methods recast the problem
through a new lens using Expectation-Maximization or Mirror Descent
Optimization, and this also results in first-order optimization with
KL divergence term in the loss function \cite{abdolmaleki2018maximum,song2019v,yang2019policy,tomar2020mirror}.

An issue with the above methods is that the previous policy used to
restrict the new policy may be suboptimal and thus unreliable in practice.
For example, due to stochasticity and approximations, the new policy
may fall into a local optimum even under trust-region optimizations.
Then in the next update, this policy will become the ``previous''
policy, and will continue pulling the next policy to stay in the local
optimum, thus slowing down the training progress. For on-policy methods
using mini-batch updates like PPO, the situation is more problematic
as the ``previous'' policy is defined as the old policy to collect
data, which can be either very far or close to the current policy.
There is no guarantee that the old policy defines a reasonable trust
region for regulating the new policy. 

In this paper, we propose a novel constrained policy iteration procedure,
dubbed Memory-Constrained Policy Optimization (MCPO), wherein a \emph{virtual
policy} representing memory of past policies regularizes each policy
update. The virtual policy forms a virtual trust region, attracting
the new policy more when the old policy performs badly, which prevents
the optimization from falling into local optimum caused by the old
policy of poor quality. As such, we measure 2 KL divergences corresponding
to the virtual and old policy, and assign different weights to the
two KL terms in building the objective function. The weights are computed
dynamically based on the performance of the two policies (the higher
performer yields higher weights). 

In contrast to prior works using heuristics (e.g. running average
or mean of past policies) to form additional trust regions \cite{wang2016sample},
we argue that the virtual policy should be determined dynamically
to maximize the performance on current training data. Thus we store
policies in a policy memory, and learn to extract the most relevant
ones to the current context. We sum the past policies in a weighted
manner wherein the weights are generated by a neural network--named
the\emph{ attention network}, which takes the information of the current,
the old and the last virtual policy as the input. The attention network
is optimized to maximize the approximate expected advantages of the
virtual policy. We jointly optimize the policy and attention networks
to train our system, alternating between sampling data from the policy
and updating the networks in a mini-batch manner. 

We verify our proposed MCPO through a diverse set of experiments and
compare our performance with that of recent constrained policy optimization
baselines. In our experiment on classical control tasks, amongst tested
models, MCPO consistently achieves better performance across tasks
and hyperparameters. Our testbed on 6 Mujoco tasks shows that MCPO
with a big policy memory is performant where the attention network
plays an important role. We also demonstrate MCPO's capability of
learning efficiently on sparse reward and high-dimensional problems
such as navigation and Atari games. Finally, our ablation study highlights
the necessity of MCPO's components such as the virtual policy and
the attention network.

%% file: bg.tex
In this section, we briefly review some fundamental constrained policy
optimization approaches. A general idea is to force the new policy
$\pi_{\theta}$ to be close to a recent policy $\pi_{\theta_{old}}$.
In this paper, we usually refer to a policy via its parameters (i.e.
policy $\theta$ means policy $\pi_{\theta}$). We also use finite-horizon
estimators for the advantage with discount factor $\gamma\in\left(0,1\right)$
and horizon $T$.

\textbf{Conservative Policy Iteration (CPI) }This method starts with
a basic objective of policy gradient algorithms, which is to maximize
the expected advantage $\hat{A}_{t}$.

\[
L^{CPI}\left(\theta\right)=\hat{\mathbb{E}}_{t}\left[\frac{\pi_{\theta}\left(a_{t}|s_{t}\right)}{\pi_{\theta_{old}}\left(a_{t}|s_{t}\right)}\hat{A}_{t}\right]
\]
where the advantage $\hat{A}_{t}$ is a function of returns collected
from $\left(s_{t},a_{t}\right)$ by using $\pi_{\theta_{old}}$ (see
Appendix \ref{subsec:The-advantage-function}) and $\hat{\mathbb{E}}_{t}\left[\cdot\right]$
indicates the empirical average over a finite batch of data. To constrain
policy updates, the new policy is a mixture of the old and the greedy
policy: $\tilde{\theta}=\mathrm{argmax\;}L^{CPI}\left(\theta\right)$.
That is, $\theta=\alpha\theta_{old}+(1-\alpha)\text{\ensuremath{\tilde{\theta}}}$
where $\alpha$ is the mixture hyperparameter \cite{Kakade2002ApproximatelyOA}.
As the data is sampled from the previous iteration\textquoteright s
policy $\theta_{old}$, the objective needs importance sampling estimation.
Hereafter, we denote $\frac{\pi_{\theta}\left(a_{t}|s_{t}\right)}{\pi_{\theta_{old}}\left(a_{t}|s_{t}\right)}$
as $\tau_{t}\left(\theta\right)$ for short.

\textbf{KL-Regularized Policy Optimization }To enforce the constraint,
one can jointly maximize the advantage and minimize KL divergence
between the new and old policy, which ensures monotonic improvement
\cite{schulman2015trust}.

\[
L^{KL}\left(\theta\right)=\hat{\mathbb{E}}_{t}\left[\tau_{t}\left(\theta\right)\hat{A}_{t}-\beta KL\left[\pi_{\theta_{old}}\left(\cdot|s_{t}\right),\pi_{\theta}\left(\cdot|s_{t}\right)\right]\right]
\]
where $\beta$ is a hyperparameter that controls the update conservativeness,
which can be fixed (KL Fixed) or changed (KL Adaptive) during training
\cite{schulman2017proximal}.

\textbf{Trust Region Policy Optimization (TRPO) }The method optimizes
the expected advantage with hard constraint \cite{schulman2015trust}.
This is claimed as a practical implementation, less conservative than
the theoretically justified algorithm using KL regularizer mentioned
above. 

\begin{align*}
L^{TRPO}\left(\theta\right) & =\hat{\mathbb{E}}_{t}\left[\tau_{t}\left(\theta\right)\hat{A}_{t}\right]\\
\mathrm{st\:\,}\delta & \geq KL\left[\pi_{\theta_{old}}\left(\cdot|s_{t}\right),\pi_{\theta}\left(\cdot|s_{t}\right)\right]
\end{align*}
where $\delta$ is the KL constraint radius. 

\textbf{Proximal Policy Optimization (PPO) }PPO is a family of constrained
policy optimization, which uses first-order optimization and mini-batch
updates including KL Adaptive and clipped PPO. In this paper, we use
PPO to refer to the method that limits the change in policy by clipping
the loss function (clipped PPO) \cite{schulman2017proximal}. The
objective $L^{PPO}$ is defined as

\[
\hat{\mathbb{E}}_{t}\left[\min\left(\tau_{t}\left(\theta\right)\hat{A}_{t},\mathrm{clip}\left(\tau_{t}\left(\theta\right),1-\epsilon,1+\epsilon\right)\hat{A}_{t}\right)\right]
\]
where $\epsilon$ is the clip hyperparameter. 

In all the above methods, $\theta$ is the currently optimized policy,
which is also referred to as the current policy. $\theta_{old}$ represents
a past policy, which can be one or many update steps before the current
policy. In either case, the rule to decide $\theta_{old}$ is fixed
throughout training. If $\theta_{old}$ is suboptimal, it is unavoidable
that the following updates will be negatively impacted. We will address
this issue in the next section.

%% file: method.tex
In trust-region policy gradient methods with mini-batch updates such
as PPO, the old policy $\theta_{old}$ is often the last \emph{``sampling''}
policy for collecting observations from the environment. This policy
is, by design, fixed during updates of the main policy until the next
interaction with the environment. This means $\theta_{old}$ may not
immediately precede the current policy $\theta$ but may be from many
(update) steps before. Since $\theta_{old}$ is not up-to-date and
could possibly be not good, using it to constrain the current policy
$\theta$ can cause the suboptimal update of $\theta$. Similarly,
the immediately preceding policy could also be poor in quality and
thus, using it as $\theta_{old}$ could be detrimental to the optimization.

To tackle this issue, we propose to learn to constrain $\theta$ towards
a \emph{weighted combination of multiple past policies} besides $\theta_{old}$.
We argue that we can optimize the attention network to ensure the
combined policy is optimal. Below is formal description of our method.

\begin{algorithm}[t]
\begin{algorithmic}[1]
\REQUIRE{A policy buffer $\mathcal{M}$, an initial policy $\pi_{\theta_{old}}$}. $T$, $K$, $B$ are the learning horizon, number of update epochs, and batch size, respectively.
\STATE{Initialize $\psi_{old} \leftarrow \theta_{old}$, $\theta \leftarrow \theta_{old}$}
\FOR{$iteration=1,2,...$}
\STATE{Run policy $\pi_{\theta_{old}}$ in environment for $T$ timesteps. Compute advantage estimates $\hat{A}_{1}$, ..., $\hat{A}_{T}$}
\FOR{$epoch=1,2,...K$}
\FOR{$batch=1,2,...T/B$}
\STATE{Compute $\psi$ (Eq. \ref{eq:new_psi}) using $\psi_{old}$, $\theta$, $\theta_{old}$,  optimize $\theta$ and $\varphi$ by maximizing $L^{MCPO}$ (Eq. \ref{eq:l_mcpo})}
\STATE{$\mathbf{if}$ $D\left(\theta,\psi\right)>D\left(\theta_{old},\psi\right)$ $\mathbf{then}$ add $\theta$ to $\mathcal{M}$}
\STATE{$\mathbf{if}$ $|\mathcal{M}|>N$  $\mathbf{then}$ remove the oldest item in $\mathcal{M}$ }
\STATE{$\psi_{old} \leftarrow \psi$}
\ENDFOR
\ENDFOR
\STATE{$\theta_{old} \leftarrow \theta$}
\ENDFOR
\end{algorithmic}

\caption{Memory-Constrained Policy Optimization.\label{alg:Memory-Constrained-Policy-Optimi}}
\end{algorithm}

\subsection{Virtual Policy}

\paragraph{Computing the virtual policy via past policies}

The virtual policy should be determined based on the past policies
and their contexts such as quality, distance or entropy. A simple
strategy such as taking average of past policies is likely suboptimal
as the quality of these policies vary and some can be irrelevant to
the current learning context. Let $\psi$ be the weighted combination
of $M$ past policies $\left\{ \theta_{i}\right\} _{i=1}^{M}$, which
we refer to as a \emph{``virtual''} policy for naming convenience,
$\psi$ is computed as follows:
\begin{equation}
\psi=\sum_{i=1}^{M}f_{\varphi}(v)_{i}\theta_{i}\label{eq:new_psi}
\end{equation}
where $f_{\varphi}$ is a neural network parameterized by $\varphi$
which outputs softmax attention weights over the $M$ past policies,
$v$ is a \emph{``context''} vector capturing different relations
among the current policy $\theta$, the last sampling policy $\theta_{old}$,
and the last virtual policy $\psi_{old}$. We build the context by
extracting specific features: pair-wise distances between policies,
the empirical returns of these policies, policy entropy and value
losses (details in Appendix Table \ref{tab:Features-of-the}). Intuitively,
these features suggest which virtual policy will yield high performance
(e.g., a virtual policy that is closer to the policy that obtained
high return and low value loss). The details of $f_{\varphi}$ training
will be given in Sec. \ref{subsec:Policy-Optimization-with}.

\paragraph{Storing  past policies with diversity-promoting\emph{ }writing}

We use a memory buffer $\mathcal{M}$ to store past policies. We treat
$\mathcal{M}$ as a \emph{queue} with maximum capacity $N$, which
means a new policy $\theta$ will be added to the end of $\mathcal{M}$
and if $\mathcal{M}$ is full, the oldest policy will be discarded.
However, we do \emph{not} add any new policy $\theta$ to $\mathcal{M}$
unconditionally but only when $\theta$ satisfies our \emph{``diversity-promoting''}
condition. Let $D\left(a,b\right)=\hat{\mathbb{E}}_{t}\left[KL\left[\pi_{a}\left(\cdot|s_{t}\right),\pi_{b}\left(\cdot|s_{t}\right)\right]\right]$
denote the ``distance'' between 2 policies $\pi_{a}$ and $\pi_{b}$,
the \emph{diversity-promoting }writing is defined as:
\begin{equation}
\text{Adding \ensuremath{\theta} to \ensuremath{\mathcal{M}} if \ensuremath{D(\theta,\psi)\geq D(\theta_{old},\psi)}}\label{eq:write_rule}
\end{equation}
where $D(\theta_{old},\psi)$ serves as a threshold. This condition
makes sure that the policy to be added is far enough from $\psi$.
It is reasonable because if $\theta$ is too similar to $\psi$ (in
regard of $\theta_{old}$), the advantage of storing multiples past
policies in order to find a good one will disappear. We will elaborate
more on this in Sec. \ref{subsec:Ablation-Study}.

\subsection{Policy Optimization with Two Trust Regions\label{subsec:Policy-Optimization-with}}

\paragraph{Optimizing the policy parameters $\theta$}

During policy optimization, we make use of both $\theta_{old}$ and
$\psi$ to constrain $\theta$. Hence, $\theta_{old}$ and $\psi$
form 2 trust regions, and we aim to enforce the new policy to be closer
to the better one. To this end, we propose to learn a new policy $\theta$
by maximizing the following objective function:
\begin{align}
L_{1}(\theta)= & \hat{\mathbb{E}}_{t}\left[\tau_{t}(\theta)\hat{A}_{t}\right]\nonumber \\
 & -\beta\hat{\mathbb{E}}_{t}\bigg[\left(1-\alpha_{t}\left(\cdot|s_{t}\right)\right)KL\left[\pi_{\theta_{old}}\left(\cdot|s_{t}\right),\pi_{\theta}\left(\cdot|s_{t}\right)\right]\nonumber \\
 & \ \ \ \ \ \ \ \ +\alpha_{t}\left(\cdot|s_{t}\right)KL\left[\pi_{\psi}\left(\cdot|s_{t}\right),\pi_{\theta}\left(\cdot|s_{t}\right)\right]\bigg]\label{eq:2kl}
\end{align}
where $\beta$ is the weight balancing between the main objective
and the policy constraints, $\alpha_{t}$ is the weight balancing
between constraining $\theta$ towards $\psi$ and $\theta$ towards
$\theta_{old}$. The expectation is estimated by taking average over
$t$ in a mini-batch of sampled data. We note that in the early stage
of learning, the attention network is not trained well, and thus the
quality of the virtual policy $\psi$ may be worse than $\theta_{old}$.
In the long-term, $\psi$ will get better and tend to provide a better
trust-region. Hence, we need to use both trust regions to ensure maximal
performance at any learning stage. We can also prove that using our
proposed two trust regions guarantees monotonic policy improvement
(see Appendix. \ref{subsec:Theoretical-analysis-of}). Below we introduce
a mechanism to automatically determine the contribution of the two
trust regions through computing $\alpha$ and $\beta$ coefficients. 

Intuitively, if the virtual policy is better than the old policy,
the new policy should be kept close to the virtual policy and vice
versa. Hence, $\alpha_{t}$ should be proportional to the contribution
of $\psi$ to the final performance with regard to that of $\theta_{old}$.
Thus, we define: $\alpha_{t}\left(\cdot|s_{t}\right)=\frac{\exp(R_{t}(\psi))}{\exp(R_{t}(\psi))+\exp(R_{t}(\theta_{old}))}$
where $R_{t}(\psi)$, $R_{t}(\theta_{old})$ are the \emph{estimated
returns} corresponding to $\psi$ and $\theta_{old}$, respectively.
We estimate $R_{t}(\cdot)$ via weighted importance sampling, that
is, $R_{t}(\cdot)=\tau_{t}(\cdot)\hat{A}_{t}$ where both are computed
using the same $s_{t}$. We also dynamically adjust $\beta$ by switching
between 2 values $\beta_{min}$ and $\beta_{max}$ ($0<\beta_{min}<\beta_{max}$)
as follows:
\begin{equation}
\beta=\begin{cases}
\beta_{max} & \text{if }\,D(\theta_{old},\theta)>D(\theta_{old},\psi)\\
\beta_{min} & \text{otherwise}
\end{cases}\label{eq:switch_rule}
\end{equation}
where $D(\cdot,\cdot)$ is again the KL ``distance''. The reason
behind this update of $\beta$ is to encourage stronger enforcement
of the constraint when $\theta$ is too far from $\theta_{old}$.
Unlike using a fixed threshold $d_{targ}$ to change $\beta$ (e.g
in KL Adaptive, if $D\left(\theta_{old},\theta\right)>d_{targ}$,
increase $\beta$ \cite{schulman2017proximal}), we make use of $\psi$
as a reference for selecting $\beta$. This allows a dynamic threshold
that varies depending on the current learning. We name this mechanism
as \emph{switching-$\beta$ }rule.

\paragraph{Optimizing the attention network parameters $\varphi$}

To encourage the attention network $f_{\varphi}$ (Eq.~\ref{eq:new_psi})
to produce the optimal attention weights over past policies, we maximize
the following objective:

\begin{equation}
L_{2}(\varphi)=\mathbb{\hat{E}}_{t}\left[R_{t}(\psi_{\varphi})\right]\label{eq:l2}
\end{equation}
 which is the approximate expected return w.r.t. $\psi_{\varphi}$.
We write $\psi_{\varphi}$ to emphasize that $\psi$ is a function
of $\varphi$ (as formulated in Eq.~\ref{eq:new_psi}). We use gradient
ascent to update $\varphi$ by backpropagating the gradient $\frac{\mathcal{\partial L}_{2}}{\partial\varphi}$
through the attention network. 

By learning $f_{\varphi}$, we can guarantee that the virtual policy
$\psi$ is the best combination of past policies in terms of return.
Note that for on-policy learning setting, optimizing ``soft'' attention
weights is usually more robust than searching for the best past policy
in $\mathcal{M}$ because the estimation of the expected return using
current data samples can be noisy and is not always reliable. Also,
searching is computationally more expensive than using the attention. 

\begin{table*}
\begin{centering}
\begin{tabular}{lccc}
\hline 
\multirow{2}{*}{{\small{}Model}} & {\small{}Pendulum} & {\small{}LunarLander} & {\small{}BWalker}\tabularnewline
 & {\small{}1M} & {\small{}1M} & {\small{}5M}\tabularnewline
\hline 
{\small{}KL Adaptive ($d_{targ}=0.01$)} & {\small{}-147.52\textpm 9.90} & \emph{\small{}254.26\textpm 19.43} & {\small{}247.70\textpm 14.16}\tabularnewline
{\small{}KL Fixed ($\beta=0.1$)} & {\small{}-464.29\textpm 426.27} & \emph{\small{}256.75\textpm 20.53} & \emph{\small{}\uline{263.56\textpm 10.04}}\tabularnewline
{\small{}PPO (clip $\epsilon=0.3$)} & {\small{}-591.31\textpm 229.32} & \emph{\small{}\uline{259.93\textpm 22.52}} & \emph{\small{}\uline{260.51\textpm 17.86}}\tabularnewline
{\small{}MDPO ($\beta_{0}=2$)} & \emph{\small{}\uline{-135.52\textpm 5.28}} & {\small{}227.76\textpm 16.96} & {\small{}226.80\textpm 15.67}\tabularnewline
{\small{}VMPO ($\alpha_{0}=1$)} & \emph{\small{}-139.50\textpm 5.54} & {\small{}212.85\textpm 43.35} & {\small{}238.82\textpm 11.11}\tabularnewline
\hline 
{\small{}MCPO ($N=5$)} & \textbf{\small{}-133.42\textpm 4.53} & \emph{\small{}\uline{262.23\textpm 12.47}} & \emph{\small{}\uline{265.80\textpm 5.55}}\tabularnewline
{\small{}MCPO ($N=10$)} & {\small{}-146.88\textpm 3.78} & \emph{\small{}\uline{263.04\textpm 11.48}} & \textbf{\small{}266.26\textpm 8.87}\tabularnewline
{\small{}MCPO ($N=40$)} & \emph{\small{}\uline{-135.57\textpm 5.22}} & \textbf{\small{}267.19\textpm 13.42} & {\small{}249.51\textpm 12.75}\tabularnewline
\hline 
\end{tabular}
\par\end{centering}
\caption{Mean and std. over 5 runs on classical control tasks (with number
of training steps). Bold denotes the best mean. Underline denotes
good results (if exist), statistically indifferent from the best in
terms of Cohen effect size less than 0.5. The baselines are reported
with best hyperparameters.\label{tab:cc}}
\end{table*}

\paragraph{Final Objective}

We train the whole system by maximizing the following objective:
\begin{equation}
L^{MCPO}=L_{1}(\theta)+L_{2}(\varphi)\label{eq:l_mcpo}
\end{equation}
where MCPO stands for \emph{Memory-Constrained Policy Optimization}.
We optimize $\theta$ and $\psi$ alternately by fixing one and learning
the other. We implement MCPO using minibatch update procedure \citep{schulman2017proximal}.
MCPO Pseudocode is given in in Algo \ref{alg:Memory-Constrained-Policy-Optimi}.
For notational simplification, the algorithm uses 1 actor.

%% file: exp.tex
In our experiments, we show our optimization scheme is superior in
different aspects: hyperparameter sensitivity, sample efficiency and
consistent performance in continuous and discrete action spaces. The
main baselines are recent on-policy constrained methods that use first-order
optimization, in which most of them employ KL terms in the objective
function. They are KL Adaptive, KL Fixed, PPO \cite{schulman2017proximal},
MDPO \cite{tomar2020mirror}, VMPO \cite{song2019v} and TRGPPO \cite{wang2019trust}
. We also include second-order methods such as TRPO \cite{schulman2015trust}
and ACKTR \cite{wu2017scalable}. Across experiments, for MCPO, we
fix $\beta_{max}=10$, $\beta_{min}=0.01$ and only tune $N$. More
details on the baselines and tasks are given in Appendix \ref{subsec:Baselines-and-tasks}.

\subsection{Classical Control}

In this section, we compare MCPO to other first-order policy gradient
methods (KL Adaptive, KL Fixed, PPO, MDPO and VMPO) on 3 classical
control tasks: Pendulum, LunarLander and BipedalWalker, which are
trained for one, one and five million environment steps, respectively.
Here, we are curious to know how the model performance fluctuates
as the hyperparameters vary. For each model, we choose one hyperparameter
that controls the conservativeness of the policy update, and we try
different values for the signature hyperparameter while keeping the
others the same. For example, for PPO, we tune the clip value $\epsilon$;
for KL Fixed we tune $\beta$ coefficient and these possible values
are chosen following prior works. For our MCPO, we tune the size of
the policy memory $N$ (5, 10 and 40). We do not try bigger policy
memory size to keep MCPO running efficiently (see Appendix \ref{subsec:Details-on-Classical}
for details).

Table \ref{tab:cc} reports the results of MCPO and 5 baselines with
the best hyperparameters. For these simple tasks, tuning the hyperparameters
often helps the model achieve at least moderate performance. However,
models like KL Adaptive and VMPO cannot reach good performance despite
being tuned. PPO shows good results on LunarLander and BipedalWalker,
yet underperforms others on Pendulum. Interestingly, if tuned properly,
the vanilla KL Fixed can show competitive results compared to PPO
and MDPO in BipedalWalker. Amongst all, our MCPO with suitable $N$
achieves the best performance on all tasks. Remarkably, its performance
does not fluctuate much as $N$ changes from 5 to 40, often obtaining
good and best results. We speculate that for simple tasks, either
with small or big $\mathcal{M}$, the policy attention can always
find optimal virtual policy $\psi$, and thus, ensures stable performance
of MCPO. On the contrary, other methods observe a clear drop in performance
as hyperparameters change (see Appendix \ref{subsec:Details-on-Classical}
for learning curves and the full table). 

\begin{figure*}
\begin{centering}
\includegraphics[width=0.9\linewidth]{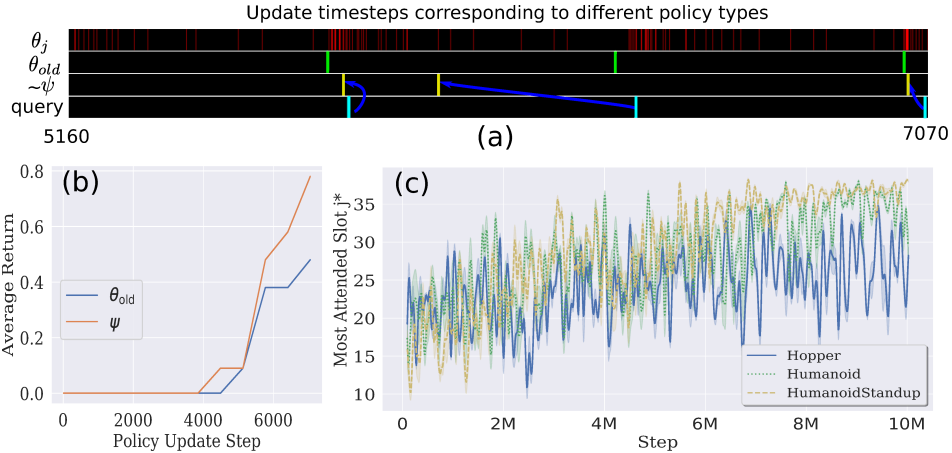}
\par\end{centering}
\caption{(a) Policy analysis on Unlock. First row (red lines): steps where
a policy is added to $\mathcal{M}$, i.e. the steps of $\theta_{j}$.
Second row (green lines): steps of old policies $\theta_{old}$. Third
row (yellow lines): steps of mostly attended policy, approximating
$\psi$. Fourth row (cyan lines): 3 steps of interest where we want
to find their attended steps. Blue arrows link a query step and the
step that receives highest attention. (b) Quality of $\psi$ vs. $\theta_{old}$.
Average return collected by $\psi$ and $\theta_{old}$ at different
stages of training. (c) 3 Mujoco tasks. The slot in $\mathcal{M}$
received the highest attention $j^{*}=\mathrm{argmax_{j}}\,f_{\varphi}\left(v_{context}\right)_{j}$
over time. \label{fig:pindex}}
\end{figure*}

\subsection{Navigation }

Here, we validate our method on sparse reward environments using MiniGrid
library \cite{gym_minigrid}. In particular, we test MCPO and other
baselines (same as above) on Unlock and UnlockPickup tasks. In these
tasks, the agent navigates through rooms and picks up objects to complete
the episode. The agent only receives reward +1 if it can complete
the episode successfully. For sample efficiency test, we train all
models on Unlock (find key and open the door) and UnlockPickup (find
key, open the door and pickup an object), for only 100,000 and 1 million
environment steps, respectively. The models use the best conservative
hyperparameters found in the previous task (more in Appendix \ref{subsec:Details-on-MiniGrid}). 

Appendix's Fig. \ref{fig:panl} shows the learning curves of examined
models on these two tasks. For Unlock task, except for MCPO and VMPO,
100,000 steps seem insufficient for other models to learn useful policies.
When trained with 1 million steps on UnlockPickup, the baselines can
find better policies, yet still underperform MCPO. Here VMPO shows
faster learning progress than MCPO at the beginning, however it fails
to converge to the best solution. Our MCPO is the best performer,
consistently ending up with average return of 0.9 (90\% of episodes
finished successfully). To achieve sample-efficiency, MCPO needs to
store and search for good policies (rarely found in sparse reward
problems), and adhere to it during optimization. MCPO's success may
be attributed to utilizing the virtual policy that has the highest
return.

To illustrate how the virtual policy supports MCPO's performance,
we analyze the relationships between the old ($\theta_{old}$), the
virtual policy ($\psi$) and the policies stored in $\mathcal{M}$
($\theta_{j}$) throughout Unlock training. Fig. \ref{fig:pindex}
(a) plots the location of these policies over a truncated period of
training (from update step 5160 to 7070). Due to diversity-promoting
rule, the steps where policies are added to $\mathcal{M}$ can be
uneven (first row-red lines), often distributed right after the locations
of the old policy (second row-green lines). We query at $10$-th step
behind the old policy (fourth row-cyan lines) to find which policy
in $\mathcal{M}$ has the highest attention (third row-yellow lines,
linked by blue arrows). As shown in Fig. \ref{fig:pindex} (a) (second
and third row), the attended policy, which mostly resembles $\psi$,
can be further or closer to the query step than the old policy depending
on the training stage. Since we let the attention network learn to
attend to the policy that maximizes the advantage of current mini-batch
data, the attended one is not necessarily the old policy.

\begin{table*}
\begin{centering}
\begin{tabular}{lcccccc}
\hline 
\multirow{1}{*}{{\footnotesize{}Model}} & {\footnotesize{}HalfCheetah} & {\footnotesize{}Walker2d} & {\footnotesize{}Hopper} & {\footnotesize{}Ant} & {\footnotesize{}Humanoid} & {\footnotesize{}HumanoidStandup}\tabularnewline
\hline 
{\footnotesize{}TRPO} & {\footnotesize{}2,811\textpm 114} & {\footnotesize{}3,966\textpm 56} & {\footnotesize{}3,159\textpm 72} & {\footnotesize{}2,438\textpm 402} & {\footnotesize{}4,576\textpm 106} & {\footnotesize{}145,143\textpm 3,702}\tabularnewline
{\footnotesize{}PPO} & {\footnotesize{}4,753\textpm 1,614} & \textbf{\footnotesize{}5,278\textpm 594} & {\footnotesize{}2,968\textpm 1,002} & {\footnotesize{}3,421\textpm 534} & {\footnotesize{}3,375\textpm 1,684} & {\footnotesize{}155,494\textpm 6,663}\tabularnewline
{\footnotesize{}MDPO} & {\footnotesize{}4,774\textpm 1,598} & {\footnotesize{}4,957\textpm 330} & {\footnotesize{}3,153\textpm 956} & {\footnotesize{}3,553\textpm 696} & {\footnotesize{}1,620\textpm 2,145} & {\footnotesize{}90,646\textpm 5,855}\tabularnewline
{\footnotesize{}TRGPPO} & {\footnotesize{}2,811\textpm 114} & {\footnotesize{}5,009\textpm 391} & {\footnotesize{}3,713\textpm 275} & \textbf{\footnotesize{}4,796\textpm 837} & \textbf{\footnotesize{}6,242\textpm 1192} & {\footnotesize{}162,185\textpm 3755}\tabularnewline
\hline 
{\footnotesize{}Mean $\psi$} & {\footnotesize{}4,942\textpm 3,095} & {\footnotesize{}5,056\textpm 842} & {\footnotesize{}3,430\textpm 259} & \emph{\footnotesize{}4,570\textpm 548} & {\footnotesize{}353\textpm 27} & {\footnotesize{}71,308\textpm 11,113}\tabularnewline
{\footnotesize{}MCPO} & \textbf{\footnotesize{}6,173\textpm 595} & \emph{\footnotesize{}5,120\textpm 588} & \textbf{\emph{\footnotesize{}3,620\textpm 252}} & {\footnotesize{}4673\textpm 249} & {\footnotesize{}4,848\textpm 711} & \textbf{\footnotesize{}195,404\textpm 32,801}\tabularnewline
\hline 
\end{tabular}
\par\end{centering}
\caption{Mean and std. over 5 runs on 6 Mujoco tasks at 10M environment steps.
\label{tab:mujoco}}
\end{table*}
The choice of the chosen virtual policy being better than the old
policy is shown in Fig. \ref{fig:pindex} (b) where we collect several
checkpoints of virtual and old policies across training and evaluate
each of them on 10 testing episodes. Here using $\psi$ to form the
second KL constraint is beneficial as the new policy is generally
pulled toward a better policy during training. That contributes to
the excellent performance of MCPO compared to other single trust-region
baselines, especially KL Fixed and Adaptive, which are very close
to MCPO in term of objective function style. 

\subsection{Mujoco}

Next, we examine MCPO and some trust-region methods from the literature
that are known to have good performance on continuous control problems:
TRPO, PPO, MDPO and TRGPPO (an improved version of PPO). To understand
the role of the attention network in MCPO, we design a variant of
MCPO ($N=40$): \emph{Mean $\psi$}, which simply constructs $\psi$
by taking average over policy parameters in $\mathcal{M}$. This baseline
represents prior heuristic ways of building trust-region from past
policies \cite{wang2016sample}. We pick 6 hard Mujoco tasks and train
each model for 10 million environment steps. We report the results
of best tuned models in Table \ref{tab:mujoco} (see Appendix \ref{subsec:Details-on-Mujoco}
for full results).

The results show that MCPO consistently achieves good performance
across 6 tasks, where it outperforms others significantly in HalfCheetah,
Hopper, and HumanoidStandup. In other tasks, MCPO is the second best,
only slightly earning less score than the best one in Walker2d and
Ant. The variant Mean $\psi$ shows reasonable performance for the
first 4 tasks, yet almost fails to learn on the last two. Thus, mean
virtual policy can perform badly. 

To understand the effectiveness of the attention network, we visualize
the attention pattern of MCPO on the last two tasks and on Hopper-a
task that Mean $\psi$ performs well on. Fig. \ref{fig:pindex} (c)
illustrates that for the first two harder tasks, MCPO gradually learns
to favor older policies in $\mathcal{M}$ $(j^{*}>35)$, which puts
more restriction on the policy change as the model converges. This
strategy seems critical for those tasks as the difference in average
return between learned $\psi$ and Mean $\psi$ is huge in these cases.
On the other hand, on Hopper, the top attended slots are just above
the middle policies in $\mathcal{M}$ $(j^{*}\sim25)$, which means
this task prefers an average restriction. As in most cases, MCPO with
big policy memory ($N=40$, more conservativeness) is beneficial.
For those tasks where MCPO does not show clear advantage, we speculate
that conservative updates are less important. 

We also visualize $\alpha_{t}$ for HalfCheetah task, and observe
an increasing trend across training steps (see Appendix Fig. \ref{fig:alphat}).
As the attention network gets trained, the virtual policy becomes
better and can complement the old policy when the latter performs
worse. Hence, on average, the weight $\alpha_{t}$ tends to be bigger
overtime. 
\begin{center}
\par\end{center}

\subsection{Atari Games}

As showcasing the robustness of our method to high-dimensional inputs,
we execute an experiment on a subset of Atari games wherein the states
are screen images and the policy and value function approximator uses
deep convolutional neural networks. We choose 9 typical games (6 were
introduced in \cite{mnih2013playing} and 3 randomly chosen) and benchmark
MCPO against PPO, ACKTR, VMPO and PRGPPO, training all models for
only 10 million environment steps. In this experiment, MCPO uses $N=10$
and other baselines' hyperparameters are selected based on the original
papers (see Appendix \ref{subsec:Details-on-Atari}). 

\begin{table}
\begin{centering}
{\small{}}%
\begin{tabular}{cccccc}
\hline 
\multirow{1}{*}{{\footnotesize{}Model}} & \multirow{1}{*}{{\footnotesize{}PPO}} & {\footnotesize{}ACKTR} & {\footnotesize{}VMPO} & {\footnotesize{}TRGPPO} & {\footnotesize{}MCPO}\tabularnewline
\hline 
{\footnotesize{}Mean} & {\footnotesize{}131.19} & {\footnotesize{}195.52} & {\footnotesize{}18.20} & {\footnotesize{}116.80} & \textbf{\footnotesize{}229.99}\tabularnewline
{\footnotesize{}Median} & {\footnotesize{}52.85} & {\footnotesize{}25.30} & {\footnotesize{}13.56} & {\footnotesize{}43.24} & \textbf{\footnotesize{}65.78}\tabularnewline
\hline 
\end{tabular}{\small\par}
\par\end{centering}
\caption{Average normalized human score over 9 games. The performance of each
run is measured by the best checkpoint during training over 10 million
frames, averaged over 5 runs. \label{tab:atarinorm-1}}
\end{table}
As seen in Table \ref{tab:atarinorm-1}, MCPO is significantly better
than other baselines in terms of both mean and median normalized human
score. The learning curves of all models are given in Appendix Fig.
\ref{fig:atari10m}. To confirm MCPO maintains the leading performance
with more training iterations, we train competitive models for 40
million frames for the first 6 games (Appendix Fig. \ref{fig:atari40m}).
The results show that MCPO still outperforms other baselines after
40M frames. 

\subsection{Ablation Study\label{subsec:Ablation-Study}}

Finally, we verify MCPO's 3 components: virtual policy (in Eq. \ref{eq:2kl}),
switching-$\beta$ (Eq. \ref{eq:switch_rule}) and diversity-promoting
rule (Eq. \ref{eq:write_rule}). We also confirm the role of choosing
the right memory size $N$ and learning to attend to the virtual policy
$\psi$. In the task BipedalWalkerHardcore (OpenAI Gym), we train
MCPO with different configurations for 50M steps. First, we tune $N$
(5,10 and 40) using the normal MCPO with all components on and find
that $N=10$ is the best. Keeping $N=10$, we ablate or replace our
component with an alternative and report the findings as follows. 

\begin{figure*}
\begin{centering}
\includegraphics[width=1\linewidth]{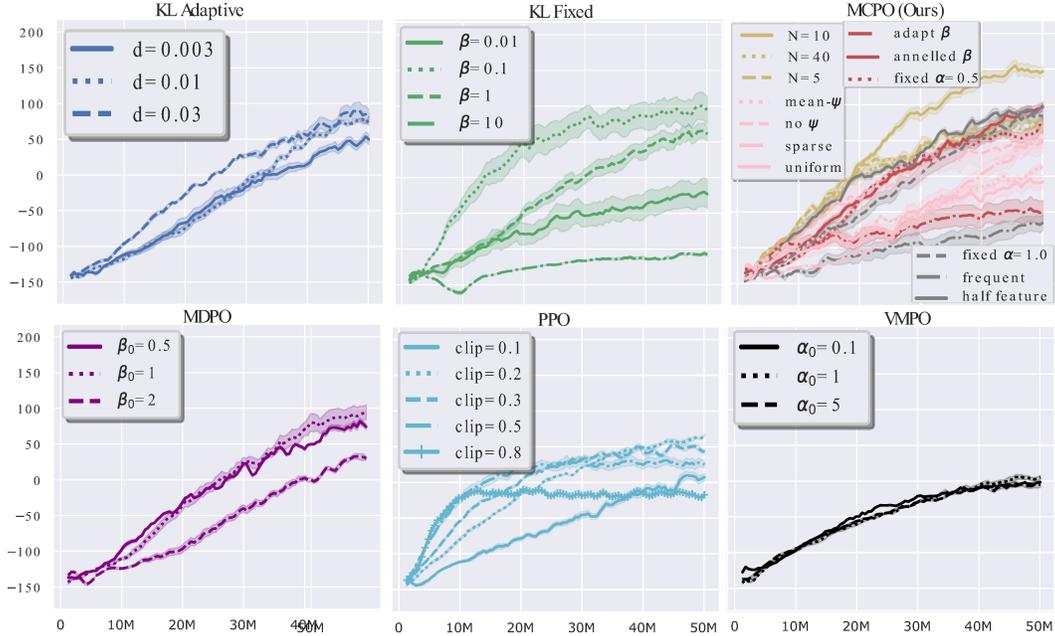}
\par\end{centering}
\caption{Ablation study on BibedalWalkerHardcore-v3: learning curves (mean
and std. over 5 runs) across 50M training steps. \label{fig:abl}}
\end{figure*}
\textbf{Virtual policy: }To show the benefit of pushing the new policy
toward the virtual policy, we implement 3 variants of MCPO (N=10)
that (i) does not use $\psi$'s KL term in Eq. \ref{eq:2kl} ($\alpha_{t}=0$),
(ii) use a fixed $\alpha_{t}=0.5$ and (iii) only use $\psi$'s KL
($\alpha_{t}=1.0$). All variants underperforms the normal MCPO by
a margin of 100 or 50 return.\textbf{ Switching-$\beta$: }The results
show that compared to the annealed $\beta$ strategy adopted from
MDPO and adaptive $\beta$ from PPO-KL, our switching-$\beta$ achieves
significantly better results with about 50 and 200 return score higher,
respectively. \textbf{Diversity-promoting writing: }We compare our
proposal with the vanilla approach that adds a new policy to $\mathcal{M}$
at every update step (frequent writing) and other versions that write
to $\mathcal{M}$ every interval of 10 and 100 update steps (uniform
and sparse writing). Frequent, uniform and sparse writing all show
slow learning progress, and ends up with low rewards. Perhaps, frequently
adding policies to $\mathcal{M}$ makes the memory content similar,
hastening the removal of older, yet maybe valuable policies. Uniform
and sparse writing are better, yet it can still add similar policies
to $\mathcal{M}$ and requires additional effort for tuning the writing
interval. \textbf{Learned $\psi$: }To benchmark, we try alternatives:
(1) using Mean $\psi$ and (2) only using half of the features in
$v_{context}$ to generate $\psi$ (Eq. \ref{eq:new_psi}). The results
confirm that the Mean $\psi$ is not a strong baseline for this task,
only reaping moderate rewards. Using less features for the context
causes information loss, hinders generations of useful $\psi$, and
thus, underperforms the full-feature version by a margin of 50 return.
For completeness, we also compare our methods to heavily tuned baselines
and confirm that the normal MCPO ($N=10$) is the best performer (see
the details in Appendix \ref{subsec:Details-on-ablation}).

%% file: related.tex
A framework for model-free reinforcement learning with policy gradient
is approximate policy iteration (API), which alternates between estimating
the advantage of the current policy, and updating the current policy's
parameters to obtain a new policy by maximizing the expected advantage
\cite{bertsekas1995neuro,sutton2000policy}. Theoretical studies have
shown that constraining policy updates is critical for API \cite{Kakade2002ApproximatelyOA,schulman2015trust,shani2020adaptive,vieillard2020momentum}.
An early attempt to improve API is Conservative Policy Iteration (CPI),
which sets the new policy as a stochastic mixture of previous greedy
policies \cite{Kakade2002ApproximatelyOA,pirotta2013safe,vieillard2020deep}.
Our paper differs from these works in three aspects: (1) we do not
directly set the new policy to the mixture, rather, we use a mixture
of previously found policies (the virtual policy) to define the trust
region constraining the new policy via KL regularization; (2) our
mixture can consist of more than 2 modes, and thus using multiple
mixture weights (attention weights); (3) we use the attention network
to learn these weights.

Also motivated by Kakade et al. (2002), TRPO extends the theory to
general stochastic policies, rather than just mixture polices, ensuring
monotonic improvement by combining maximizing the approximate expected
advantage with minimizing the KL divergence between two consecutive
policies \cite{schulman2015trust}. Arguing that optimizing this way
is too conservative and hard to tune, the authors reformulate the
objective as a constrained optimization problem to solve it with conjugate
gradient and line search. To simplify the implementation of TRPO,
Schulman et al. (2017) introduces first-order optimization methods
and code-level improvement, which results in PPO--an API method that
optimizes a clipped surrogate objective using minibatch updates. 

Constrained policy improvement can be seen as Expectation-Maximization
algorithms where minimizing the KL-term corresponds to the Expectation
step, which can be off-policy \cite{abdolmaleki2018maximum} or on-policy
\cite{song2019v}. From the mirror descent perspective, several works
also use KL divergence to regularize policy updates \cite{yang2019policy,tomar2020mirror,shani2020adaptive}.
A recent analysis also points out the advantages of using KL term
as a regularizer over a hard constraint \cite{lazic2021optimization}.
Some other works improve PPO with adaptive clip range \cite{wang2019trust}
or off-policy data \cite{fakoor2020p3o}. We instead advocate using
2 trust regions and apply our idea to a weaker backbone: PPO-penalty.
Our method is also different from \cite{fakoor2020p3o} as we do not
use off-policy data/gradients to update the policy. Overall, our approach
shares similarities with them where we also jointly optimize the approximate
expected advantage and KL constraint terms for multiple epochs of
minibatch updates. However, we propose a novel dynamic virtual policy
to construct the second trust region as a supplement to the traditional
trust region defined by the old or previous policy. 

Prior works have promoted different utilization of memory to assist
reinforcement learning. Unlike episodic memory concepts that stores
observations across agent's life \cite{blundell2016model,le2021modelbased},
the memory buffer used in this paper stores the weight parameters
of the policy network and attention mechanism is used to query this
memory. The proposed memory instead can be viewed as an instance of
program memory \cite{le2019neural,le2020neurocoder}. That said, our
memory is different from these prior works since our attention is
learned through an auxiliary objective that optimizes the quality
of the attended policy (program) with novel diversity-promoting writing
mechanisms.

%% file: discuss.tex
We have presented Memory-Constrained Policy Optimization, a new method
to regularize each policy update with two-trust regions with respect
to one single old policy and another virtual policy representing multiple
past policies. The new policy is encouraged to stay closer to the
region surrounding the policy that performs better. The virtual policy
is determined online through a learned attention to a memory of past
policies. Compared to other trust-region optimizations, MCPO shows
better performance in many environments without much hyperparameter
tuning. 

\paragraph{Limitations}

Our method introduces several new components and hyperparameters such
as $N$ and $\beta$. Due to compute limit, we have not tuned $\beta$
extensively and thus the reported results may not be the best performance
that our model can achieve. Although we find the default values $\beta_{max}=10$,
$\beta_{min}=0.01$ work well across all experiments in this paper,
we recommend adjustments if users apply our method to novel domains. 

\paragraph{Negative Societal Impacts}

Our work aims to improve optimization in RL to reduce the sample complexity
of RL algorithms. This aim is genuine, and we do not think there are
immediate harmful consequences. However, we are aware of potential
problems such as unsafe exploration committed by the agent (e.g. causing
accidents) in real-world environments (e.g. self-driving cars).  Finally,
malicious users can misuse our method for unethical purposes, such
as training harmful RL agents and robots. This issue is typical for
any machine learning algorithm, and we will do our best to prevent
it from our end.

%% file: app.tex
\renewcommand\thesubsection{\Alph{subsection}}

\subsection{Method Details\label{subsec:Method-Details}}

\subsubsection{The attention network\label{subsec:The-attention-network}}

The attention network is implemented as a feedforward neural network
with one hidden layer:
\begin{itemize}
\item Input layer: 12 units
\item Hidden layer: $N$ units coupled with a dropout layer $p=0.5$
\item Output layer: $N$ units, $\mathrm{softmax}$ activation function
\end{itemize}
$N$ is the capacity of policy memory. The 12 features of the input
$v_{context}$ is listed in Table \ref{tab:Features-of-the}. 

Now we explain the motivation behind these feature design. From these
three policies, we tried to extract all possible information. The
information should be cheap to extract and dependent on the current
data, so we prefer features extracted from the outputs of these policies
(value, entropy, distance, return, etc.). Intuitively, the most important
features should be the empirical returns, values associated with each
policy and the distances, which gives a good hint of which virtual
policy will yield high performance (e.g., a virtual policy that is
closer to the policy that obtained high return and low value loss).

\subsubsection{The advantage function\label{subsec:The-advantage-function}}

In this paper, we use GAE \cite{schulman2015high} as the advantage
function for all models and experiments

\begin{align*}
\hat{A}_{t} & =\frac{1}{N_{actor}}\sum_{i}^{N_{actor}}\sum_{k=0}^{T-t-1}\left(\gamma\lambda\right)^{k}\left(V_{target}-V\left(s_{t+k}^{i}\right)\right)
\end{align*}
where $\gamma$ is the discounted factor and $N_{actor}$ is the number
of actors. $V_{target}=r_{t+k}^{i}+\gamma V\left(s_{t+k+1}^{i}\right)$.
Note that Algo. \ref{alg:Memory-Constrained-Policy-Optimi} illustrates
the procedure for 1 actor. In practice, we use $N_{actor}$ depending
on the tasks.

\subsubsection{The objective function}

Following \cite{schulman2017proximal}, our objective function also
includes value loss and entropy terms. This is applied to all of the
baselines. For example, the complete objective function for MCPO reads

\begin{align*}
L & =L^{MCPO}-c_{1}\hat{\mathbb{E}}_{t}\left(V_{\theta}\left(s_{t}\right)-V_{target}\left(s_{t}\right)\right)^{2}\\
 & +c_{2}\hat{\mathbb{E}}_{t}\left[-\log\left(\pi_{\theta}\left(\cdot|s_{t}\right)\right)\right]
\end{align*}
where $c_{1}$ and $c_{2}$ are value and entropy coefficient hyperparameters,
respectively. $V_{\theta}$ is the value network, also parameterized
with $\theta$.

\subsection{Experimental Details\label{subsec:Experimental-Details}}

\subsubsection{Baselines and tasks\label{subsec:Baselines-and-tasks}}

All baselines in this paper share the same setting of policy and value
networks. Except for TRPO, all other baselines use minibatch training.
The only difference is the objective function, which revolves around
KL and advantage terms. We train all models with Adam optimizer. We
summarize the policy and value network architecture in Table \ref{tab:PGs-used-in}. 

The baselines ACKTR, PPO\footnote{\url{https://github.com/ikostrikov/pytorch-a2c-ppo-acktr-gail}},
TRPO\footnote{\url{https://github.com/ikostrikov/pytorch-trpo}} use
available public code (Apache or MIT License). They are Pytorch reimplementation
of OpenAI's stable baselines, which can reproduce the original performance
relatively well. For MDPO, we refer to the authors' source code\footnote{\url{https://github.com/manantomar/Mirror-Descent-Policy-Optimization}}
to reimplement the method. For VMPO, we refer to this open source
code\footnote{\url{https://github.com/YYCAAA/V-MPO_Lunarlander}}
to reimplement the method. We implement KL Fixed and KL Adaptive,
using objective function defined in Sec. \ref{sec:Background:-Constrained-Policy}.

We use environments from Open AI gyms \footnote{\url{https://gym.openai.com/envs/}},
which are public and using The MIT License. Mujoco environments use
Mujoco software\footnote{\url{https://www.roboti.us/license.html}}
(our license is academic lab). Table \ref{tab:Tasks-used-in} lists
all the environments. 

\begin{table*}
\begin{centering}
\begin{tabular}{ccc}
\hline 
Dimension & Feature & Meaning\tabularnewline
\hline 
1 & $D\left(\theta,\psi_{old}\right)$ & ``Distance'' between $\theta$ and $\psi_{old}$\tabularnewline
2 & $D\left(\theta_{old},\psi_{old}\right)$ & ``Distance'' between $\theta_{old}$ and $\psi_{old}$\tabularnewline
3 & $D\left(\theta_{old},\theta\right)$ & ``Distance'' between $\theta_{old}$ and $\theta$\tabularnewline
4 & $\hat{\mathbb{E}}_{t}\left[R_{t}\left(\psi_{old}\right)\right]$ & Approximate expected advantage of $\psi_{old}$\tabularnewline
5 & $\hat{\mathbb{E}}_{t}\left[R_{t}\left(\theta_{old}\right)\right]$ & Approximate expected advantage of $\theta_{old}$\tabularnewline
6 & $\hat{\mathbb{E}}_{t}\left[R_{t}\left(\theta\right)\right]$ & Approximate expected advantage of $\theta$\tabularnewline
7 & $\hat{\mathbb{E}}_{t}\left[-\log\left(\pi_{\psi_{old}}\left(\cdot|s_{t}\right)\right)\right]$ & Approximate entropy of $\psi_{old}$\tabularnewline
8 & $\hat{\mathbb{E}}_{t}\left[-\log\left(\pi_{\theta_{old}}\left(\cdot|s_{t}\right)\right)\right]$ & Approximate entropy of $\theta_{old}$\tabularnewline
9 & $\hat{\mathbb{E}}_{t}\left[-\log\left(\pi_{\theta}\left(\cdot|s_{t}\right)\right)\right]$ & Approximate entropy of $\theta$\tabularnewline
10 & $\hat{\mathbb{E}}_{t}\left(V_{\psi_{old}}\left(s_{t}\right)-V_{target}\left(s_{t}\right)\right)^{2}$ & Value loss of $\psi_{old}$\tabularnewline
11 & $\hat{\mathbb{E}}_{t}\left(V_{\theta_{old}}\left(s_{t}\right)-V_{target}\left(s_{t}\right)\right)^{2}$ & Value loss of $\theta_{old}$\tabularnewline
12 & $\hat{\mathbb{E}}_{t}\left(V_{\theta}\left(s_{t}\right)-V_{target}\left(s_{t}\right)\right)^{2}$ & Value loss of $\theta$\tabularnewline
\hline 
\end{tabular}
\par\end{centering}
\caption{Features of the context vector.\label{tab:Features-of-the}}
\end{table*}
\begin{table*}
\begin{centering}
\begin{tabular}{cc}
\hline 
Input type & \textbf{Policy/Value networks}\tabularnewline
\hline 
\multirow{2}{*}{Vector} & 2-layer feedforward\tabularnewline
 & net (tanh, h=64)\tabularnewline
\hline 
\multirow{3}{*}{Image} & 3-layer ReLU CNN with\tabularnewline
 & kernels $\left\{ 32/8/4,64/4/2,32/3/1\right\} $+2-layer\tabularnewline
 & feedforward net (ReLU, h=512)\tabularnewline
\hline 
\end{tabular}
\par\end{centering}
\caption{Network architecture shared across baselines. \label{tab:PGs-used-in}}
\end{table*}
\begin{table*}
\begin{centering}
\begin{tabular}{ccc}
\hline 
\multirow{2}{*}{\textbf{Tasks}} & \textbf{Continuous} & \textbf{Gym}\tabularnewline
 & \textbf{action} & \textbf{category}\tabularnewline
\hline 
Pendulum-v0 & \multirow{2}{*}{$\mathit{\mathtt{X}}$} & Classical\tabularnewline
LunarLander-v2 &  & \multirow{2}{*}{Box2d}\tabularnewline
BipedalWalker-v3 & $\checked$ & \tabularnewline
\hline 
Unlock-v0 & \multirow{2}{*}{$\mathit{\mathtt{X}}$} & \multirow{2}{*}{MiniGrid}\tabularnewline
UnlockPickup-v0 &  & \tabularnewline
\hline 
MuJoCo tasks (v2): HalfCheetah & \multirow{3}{*}{$\checked$} & \multirow{3}{*}{MuJoCo}\tabularnewline
Walker2d, Hopper, Ant &  & \tabularnewline
Humanoid, HumanoidStandup &  & \tabularnewline
\hline 
Atari games (NoFramskip-v4): & \multirow{4}{*}{$\mathit{\mathtt{X}}$} & \multirow{4}{*}{Atari}\tabularnewline
Beamrider, Breakout &  & \tabularnewline
Enduro, Gopher  &  & \tabularnewline
Seaquest, SpaceInvaders &  & \tabularnewline
\hline 
BipedalWalkerHardcore-v3 & $\checked$ & Box2d\tabularnewline
\hline 
\end{tabular}
\par\end{centering}
\caption{Tasks used in the paper. \label{tab:Tasks-used-in}}
\end{table*}

\subsubsection{Details on Classical Control\label{subsec:Details-on-Classical}}

\begin{table*}
\begin{centering}
\begin{tabular}{lccc}
\hline 
\multirow{2}{*}{Model} & Pendulum & LunarLander & BWalker\tabularnewline
 & 1M & 1M & 5M\tabularnewline
\hline 
KL Adaptive ($d_{targ}=0.003$) & -407.74\textpm 484.16 & 238.30\textpm 34.07 & 206.99\textpm 5.34\tabularnewline
KL Adaptive ($d_{targ}=0.01$) & -147.52\textpm 9.90 & \emph{254.26\textpm 19.43} & 247.70\textpm 14.16\tabularnewline
KL Adaptive ($d_{targ}=0.03$) & -601.09\textpm 273.18 & 246.93\textpm 12.57 & \emph{259.80\textpm 6.33}\tabularnewline
KL Fixed ($\beta=0.01$) & -1051.14\textpm 158.81 & 247.61\textpm 19.79 & 221.55\textpm 38.64\tabularnewline
KL Fixed ($\beta=0.1$) & -464.29\textpm 426.27 & \emph{256.75\textpm 20.53} & \emph{\uline{263.56\textpm 10.04}}\tabularnewline
KL Fixed ($\beta=1$) & \emph{-136.40\textpm 4.49} & 192.62\textpm 32.97 & 215.13\textpm 13.29\tabularnewline
PPO (clip $\epsilon=0.1$) & -282.20\textpm 243.42 & 242.98\textpm 13.50 & 205.07\textpm 19.13\tabularnewline
PPO (clip $\epsilon=0.2$) & -514.28\textpm 385.34 & \emph{256.88\textpm 20.33} & 253.58\textpm 7.49\tabularnewline
PPO (clip $\epsilon=0.3$) & -591.31\textpm 229.32 & \emph{\uline{259.93\textpm 22.52}} & \emph{\uline{260.51\textpm 17.86}}\tabularnewline
MDPO ($\beta_{0}=0.5$) & \emph{-136.45\textpm 8.21} & 247.96\textpm 4.74 & 251.18\textpm 29.10\tabularnewline
MDPO ($\beta_{0}=1$) & \emph{-139.14\textpm 10.32} & 207.96\textpm 43.86 & 245.27\textpm 10.47\tabularnewline
MDPO ($\beta_{0}=2$) & \emph{\uline{-135.52\textpm 5.28}} & 227.76\textpm 16.96 & 226.80\textpm 15.67\tabularnewline
VMPO ($\alpha_{0}=0.1$) & \emph{-144.51\textpm 7.04} & 201.87\textpm 29.48 & 236.57\textpm 10.62\tabularnewline
VMPO ($\alpha_{0}=1$) & \emph{-139.50\textpm 5.54} & 212.85\textpm 43.35 & 238.82\textpm 11.11\tabularnewline
VMPO ($\alpha_{0}=5$) & \emph{-296.48\textpm 213.06} & 222.13\textpm 35.55 & 164.40\textpm 40.36\tabularnewline
\hline 
MCPO ($N=5$) & \textbf{-133.42\textpm 4.53} & \emph{\uline{262.23\textpm 12.47}} & \emph{\uline{265.80\textpm 5.55}}\tabularnewline
MCPO ($N=10$) & -146.88\textpm 3.78 & \emph{\uline{263.04\textpm 11.48}} & \textbf{266.26\textpm 8.87}\tabularnewline
MCPO ($N=40$) & \emph{\uline{-135.57\textpm 5.22}} & \textbf{267.19\textpm 13.42} & 249.51\textpm 12.75\tabularnewline
\hline 
\end{tabular}
\par\end{centering}
\caption{Mean and std. over 5 runs on classical control tasks (with number
of training environment steps). Bold denotes the best mean. Underline
denotes good results (if exist), statistically indifferent from the
best in terms of Cohen effect size less than 0.5.\label{tab:cc-1}}
\end{table*}
For these tasks, all models share hyperparameters listed in Table
\ref{tab:cc_hp}. Besides, each method has its own set of additional
hyperparameters. For example, PPO, KL Fixed and KL Adaptive have $\epsilon$,
$\beta$ and $d_{targ}$, respectively. These hyperparameters directly
control the conservativeness of the policy update for each method.
For MDPO, $\beta$ is automatically reduced overtime through an annealing
process from 1 to 0 and thus should not be considered as a hyperparameter.
However, we can still control the conservativeness if $\beta$ is
annealed from a different value $\beta_{0}$ rather 1. We realize
that tuning $\beta_{0}$ helped MDPO (Table \ref{tab:cc-1}). We quickly
tried with several values $\beta_{0}$ ranging from 0.01 to 10 on
Pendulum, and realize that only $\beta_{0}\in\left\{ 0.5,1,2\right\} $
gave reasonable results. Thus, we only tuned MDPO with these $\beta_{0}$
in other tasks. For VMPO there are many other hyperparameters such
as $\text{\ensuremath{\eta_{0}}}$, $\alpha_{0}$, $\epsilon_{\eta}$
and $\epsilon_{\alpha}$. Due to limited compute, we do not tune all
of them. Rather, we only tune $\alpha_{0}$-the initial value of the
Lagrange multiplier that scale the KL term in the objective function.
We refer to the paper's and the code repository's default values of
$\alpha_{0}$ to determine possible values $\alpha_{0}\in\left\{ 0.1,1,5\right\} $.
For our MCPO, we can tune several hyperparameters such as $N$, $\beta_{min}$,
and $\beta_{max}$. However, for simplicity, we only tune $N\in\left\{ 5,10,40\right\} $
and fix $\beta_{min}=0.01$ and $\beta_{max}=10$. 

On our machines using 1 GPU Tesla V100-SXM2, we measure the running
time of MCPO with different $N$ compared to PPO on Pendulum task,
which is reported in Table \ref{tab:Computing-cost-of}. As $N$ increases,
the running speed of MCPO decreases. For this reason, we do not test
with $N>40$. However, we realize that with $N=5$ or $N=10$, MCPO
only runs slightly slower than PPO. We also realize that the speed
gap is even reduced when we increase the number of actors $N_{actor}$
as in other experiments. In terms of memory usage, maintaining a policy
memory will definitely cost more. However, as our policy, value and
attention networks are very simple. The maximum storage even for $N=40$
is less than 5GB. 

\begin{figure*}
\begin{centering}
\includegraphics[width=1\linewidth]{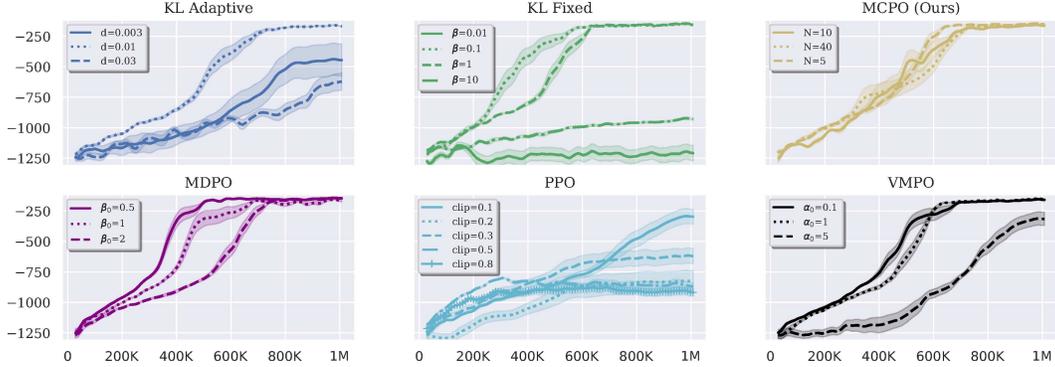}
\par\end{centering}
\caption{Pendulum-v0: learning curves (mean and std. over 5 runs) across 1M
training steps.\label{fig:Pendulum-v0:-learning-curves}}

\end{figure*}

\begin{figure*}
\begin{centering}
\includegraphics[width=1\linewidth]{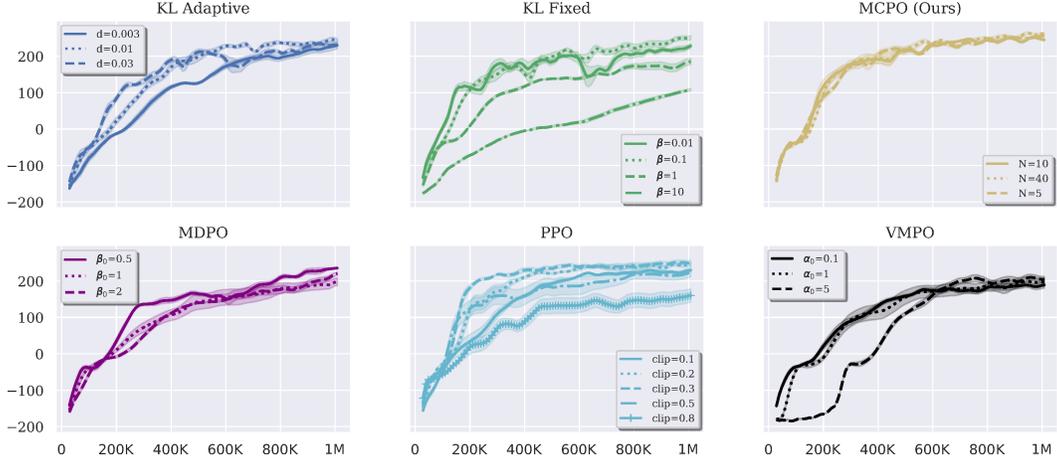}
\par\end{centering}
\caption{LunarLander-v2: learning curves (mean and std. over 5 runs) across
5M training steps.\label{fig:LunarLander-v2:-learning-curves}}
\end{figure*}

\begin{figure*}
\begin{centering}
\includegraphics[width=1\linewidth]{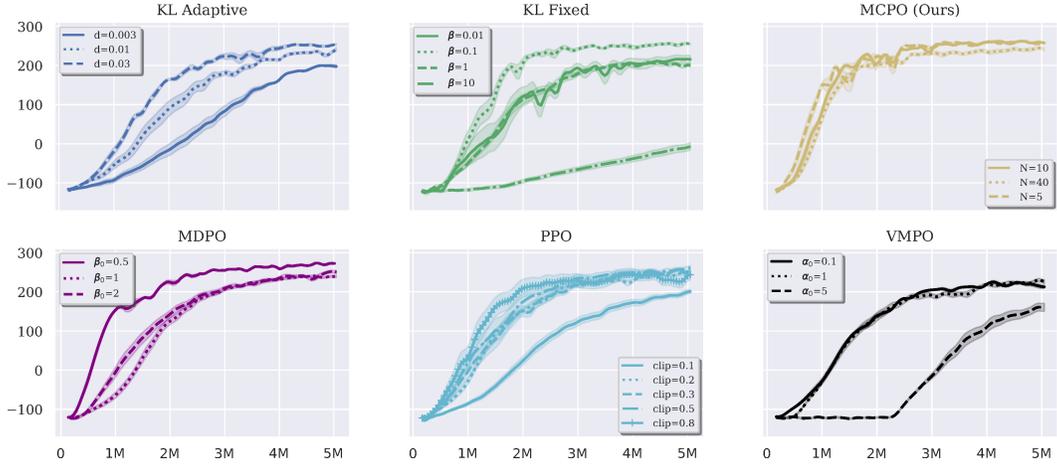}
\par\end{centering}
\caption{BipedalWalker-v3: learning curves (mean and std. over 5 runs) across
1M training steps.\label{fig:BipedalWalker-v3:-learning-curve}}
\end{figure*}

In addition to the configurations reported in Table \ref{tab:cc},
for KL Fixed and PPO, we also tested with extreme values $\beta=10$
and $\epsilon\in\left\{ 0.5,0.8\right\} $. Figs. \ref{fig:Pendulum-v0:-learning-curves},
\ref{fig:LunarLander-v2:-learning-curves} and \ref{fig:BipedalWalker-v3:-learning-curve}
visualize the learning curves of all configurations for all models. 

\begin{table*}
\begin{centering}
\begin{tabular}{cccccc}
\hline 
\multirow{2}{*}{Hyperparameter} & \multirow{2}{*}{Pendulum} & \multirow{2}{*}{LunarLander} & \multirow{2}{*}{BipedalWalker} & \multirow{2}{*}{MiniGrid} & BipedalWaker\tabularnewline
 &  &  &  &  & Hardcore\tabularnewline
\hline 
Horizon $T$ & 2048 & 2048 & 2048 & 2048 & 2048\tabularnewline
Adam step size & $3\times10^{-4}$ & $3\times10^{-4}$ & $3\times10^{-4}$ & $3\times10^{-4}$ & $3\times10^{-4}$\tabularnewline
Num. epochs $K$ & 10 & 10 & 10 & 10 & 10\tabularnewline
Minibatch size $B$ & 64 & 64 & 64 & 64 & 64\tabularnewline
Discount $\gamma$ & 0.99 & 0.99 & 0.99 & 0.99 & 0.99\tabularnewline
GAE $\lambda$ & 0.95 & 0.95 & 0.95 & 0.95 & 0.95\tabularnewline
Num. actors $N_{actor}$ & 4 & 4 & 32 & 4 & 128\tabularnewline
Value coefficient $c_{1}$ & 0.5 & 0.5 & 0.5 & 0.5 & 0.5\tabularnewline
Entropy coefficient $c_{2}$ & 0 & 0 & 0 & 0 & 0\tabularnewline
\hline 
\end{tabular}
\par\end{centering}
\caption{Network architecture shared across baselines on Pendulum, LunarLander,
BipedalWalker, MiniGrid and BipedalWaker Hardcore \label{tab:cc_hp}}
\end{table*}
\begin{table}
\begin{centering}
\begin{tabular}{cc}
\hline 
\multirow{1}{*}{Model} & \multicolumn{1}{c}{Speed (env. steps/s)}\tabularnewline
\hline 
MCPO (N=5) & 1,170\tabularnewline
MCPO (N=10) & 927\tabularnewline
MCPO (N=40) & 560\tabularnewline
PPO & 1,250\tabularnewline
\hline 
\end{tabular}
\par\end{centering}
\caption{Computing cost of MCPO and PPO on Pendulum. \label{tab:Computing-cost-of}}
\end{table}

\subsubsection{Details on MiniGrid Navigation\label{subsec:Details-on-MiniGrid}}

\begin{figure*}
\begin{centering}
\includegraphics[width=0.9\linewidth]{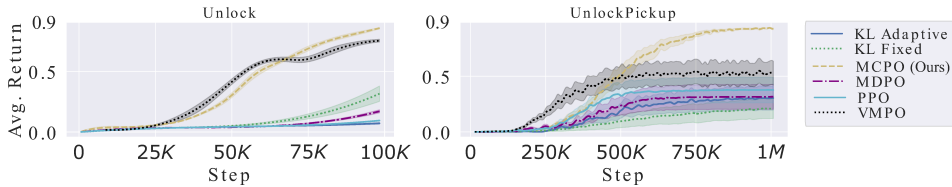}
\par\end{centering}
\caption{Unlock (left) and UnlockPickup (right)'s learning curves (mean and
std. over 10 runs). \label{fig:panl}}
\end{figure*}
Based on the results from the above tasks, we pick the best signature
hyperparameters for the models to use in this task as in Table \ref{tab:mghp}.
In particular, for each model, we rank the hyperparameters per task
(higher rank is better), and choose the one that has the maximum total
rank. For hyperparameters that share the same total rank, we prefer
the middle value. The other hyperparameters for this task is listed
in Table \ref{tab:cc_hp}.

\begin{table}
\begin{centering}
\begin{tabular}{cc}
\hline 
\multirow{1}{*}{Model} & \multicolumn{1}{c}{Chosen hyperparameter}\tabularnewline
\hline 
KL Adaptive & $d_{targ}=0.01$\tabularnewline
KL Fixed & $\beta=0.1$\tabularnewline
PPO & $\epsilon=0.2$\tabularnewline
MDPO & $\beta_{0}=0.5$\tabularnewline
VMPO & $\alpha_{0}=1$\tabularnewline
\hline 
MCPO & $N=10$\tabularnewline
\hline 
\end{tabular}
\par\end{centering}
\caption{Signature hyperparameters used in MiniGrid tasks. \label{tab:mghp}}
\end{table}

\subsubsection{Details on Mujoco\label{subsec:Details-on-Mujoco}}

For shared hyperparameters, we use the values suggested in the PPO's
paper, except for the number of actors, which we increase to $16$
for faster training as our models are trained for 10M environment
steps (see Table \ref{tab:cc_hpmj}). 

\begin{figure*}
\begin{centering}
\includegraphics[width=1\linewidth]{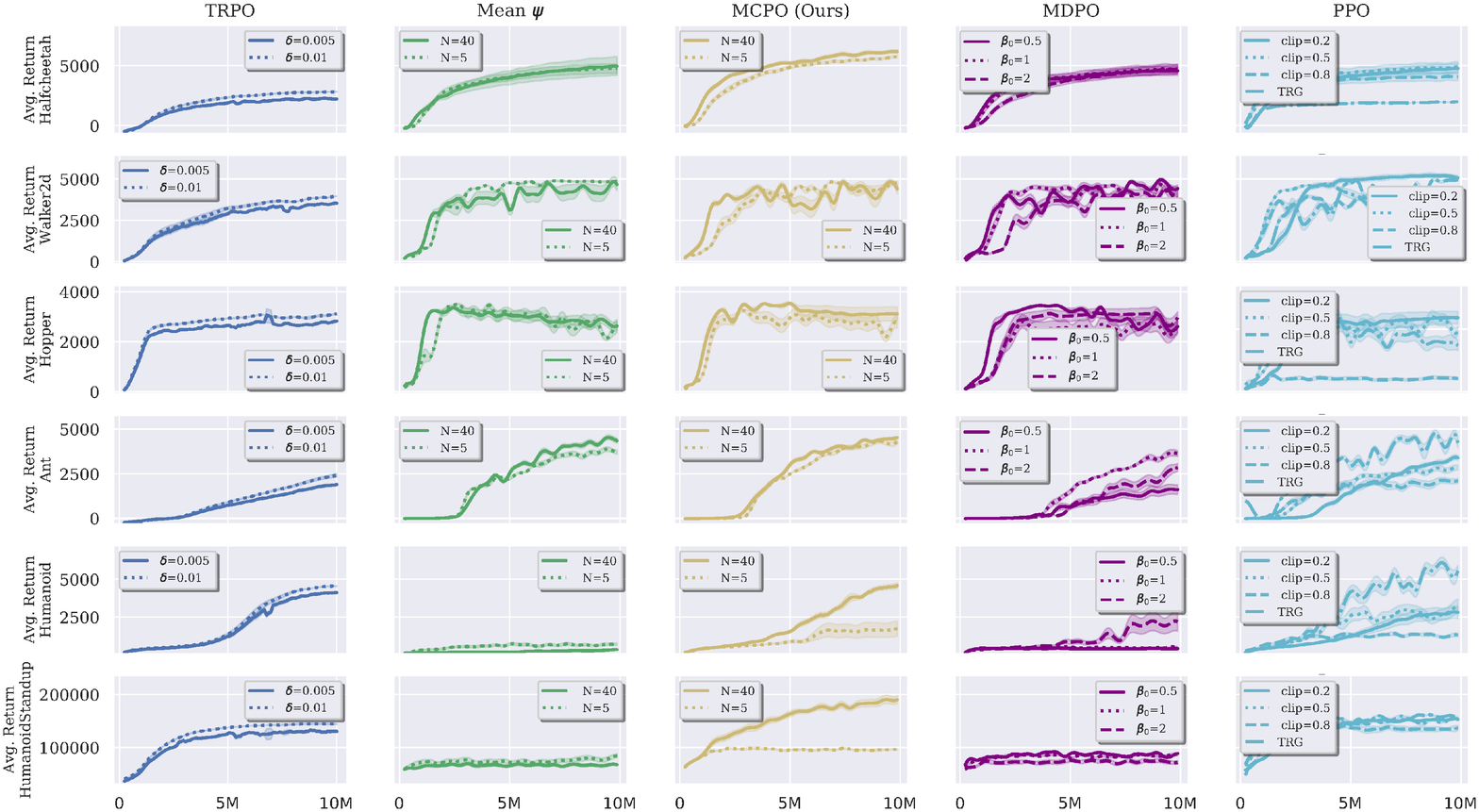}
\par\end{centering}
\caption{Mujoco: learning curves (mean and std. over 5 runs) across 10M training
steps.\label{fig:Mujoco:-learning-curves}}
\end{figure*}

\begin{figure*}
\begin{centering}
\includegraphics[width=1\linewidth]{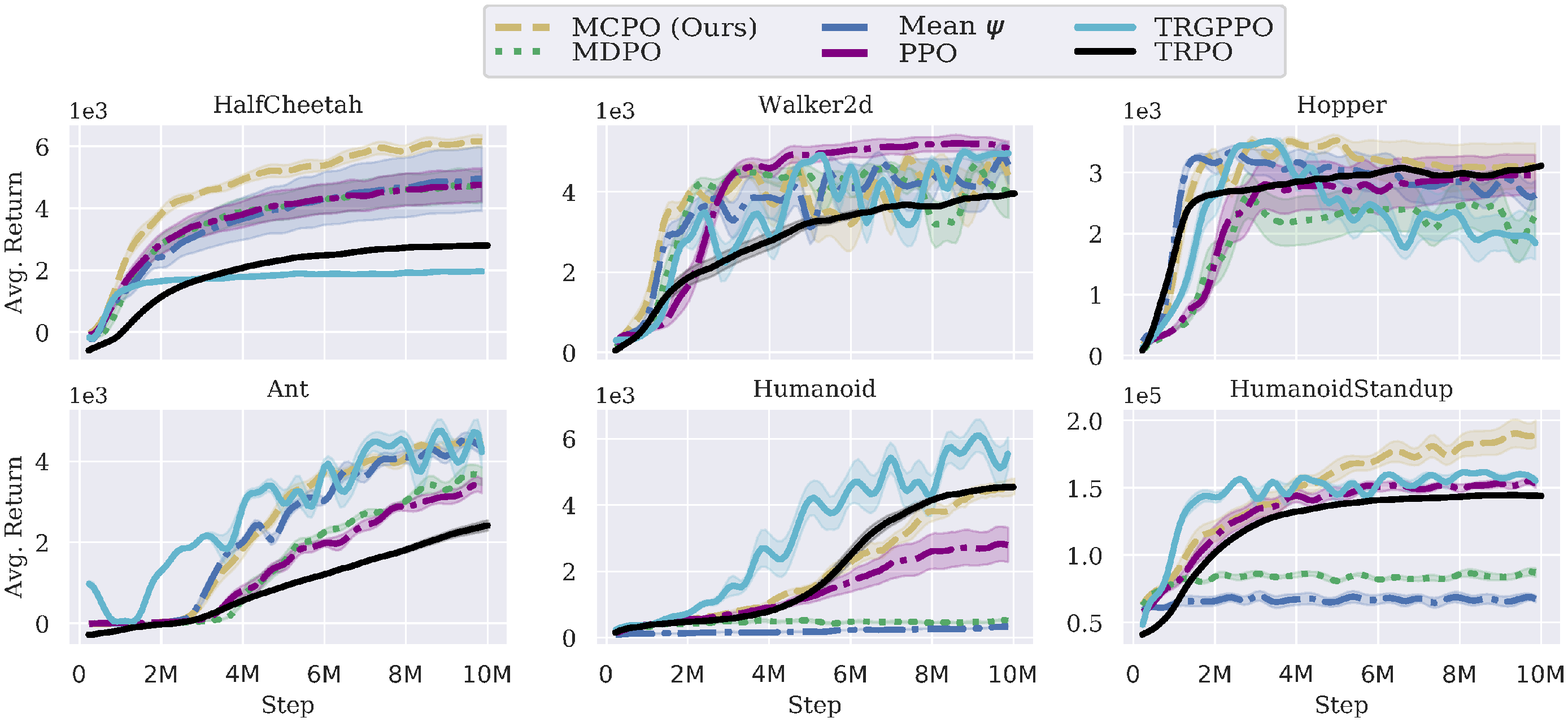}
\par\end{centering}
\caption{Mujoco: learning curves (mean and std. over 5 runs) across 10M training
steps.\label{fig:Mujoco:-learning-curves-1}}
\end{figure*}

For the signature hyperparameter of each method, we select some of
the reasonable values. For PPO, the authors already examined with
$\epsilon\in\left\{ 0.1,0.2,0.3\right\} $ on the same task and found
0.2 the best. This is somehow backed up in our previous experiments
where we did not see major difference in performance between these
values. Hence, seeking for other $\epsilon$ rather than the optimal
$\epsilon=0.2$, we ran our PPO implementation with $\epsilon\in\left\{ 0.2,0.5,0.8\right\} $.
For TRPO, the authors only used the KL radius threshold $\delta=0.01$,
which may be already the optimal hyperparameter. Hence, we only tried
$\delta\in\left\{ 0.005,0.01\right\} $. The results showed that $\delta=0.005$
always performed worse. For MCPO and Mean $\psi$, we only ran with
extreme $N\in\left\{ 5,40\right\} $. For MDPO, we still tested with
$\beta_{0}\in\left\{ 0.5,1,2\right\} $. Full learning curves with
different hyperparameter are reported in Fig. \ref{fig:Mujoco:-learning-curves}.
Learning curves including TRGPPO\footnote{We use the authors' source code \url{https://github.com/wangyuhuix/TRGPPO}
using default configuration. Training setting is adjusted to follow
the common setting as for other baselines (see Table \ref{tab:cc_hpmj}).} are reported in Fig. \ref{fig:Mujoco:-learning-curves-1}

\begin{table}
\begin{centering}
\begin{tabular}{ccc}
\hline 
\multirow{2}{*}{Hyperparameter} & \multirow{2}{*}{Mujoco} & \multirow{2}{*}{Atari}\tabularnewline
 &  & \tabularnewline
\hline 
Horizon $T$ & 2048 & 128\tabularnewline
Adam step size & $3\times10^{-4}$ & $2.5\times10^{-4}$\tabularnewline
Num. epochs $K$ & 10 & 4\tabularnewline
Minibatch size $B$ & 32 & 32\tabularnewline
Discount $\gamma$ & 0.99 & 0.99\tabularnewline
GAE $\lambda$ & 0.95 & 0.95\tabularnewline
Num. actors $N_{actor}$ & 16 & 32\tabularnewline
Value coefficient $c_{1}$ & 0.5 & 1.0\tabularnewline
Entropy coefficient $c_{2}$ & 0 & 0.01\tabularnewline
\hline 
\end{tabular}
\par\end{centering}
\caption{Network architecture shared across baselines on Mujoco and Atari \label{tab:cc_hpmj}}
\end{table}

\subsubsection{Details on Atari\label{subsec:Details-on-Atari}}

For shared hyperparameters, we use the values suggested in the PPO's
paper, except for the number of actors, which we increase to $32$
for faster training (see Table \ref{tab:cc_hpmj}). For the signature
hyperparameter of the baselines, we used the recommended value in
the original papers. For MCPO, we use $N=10$ to balance between running
time and performance. Table \ref{tab:atarihp-1} shows the values
of these hyperparameters. 

\begin{figure*}
\begin{centering}
\includegraphics[width=1\linewidth]{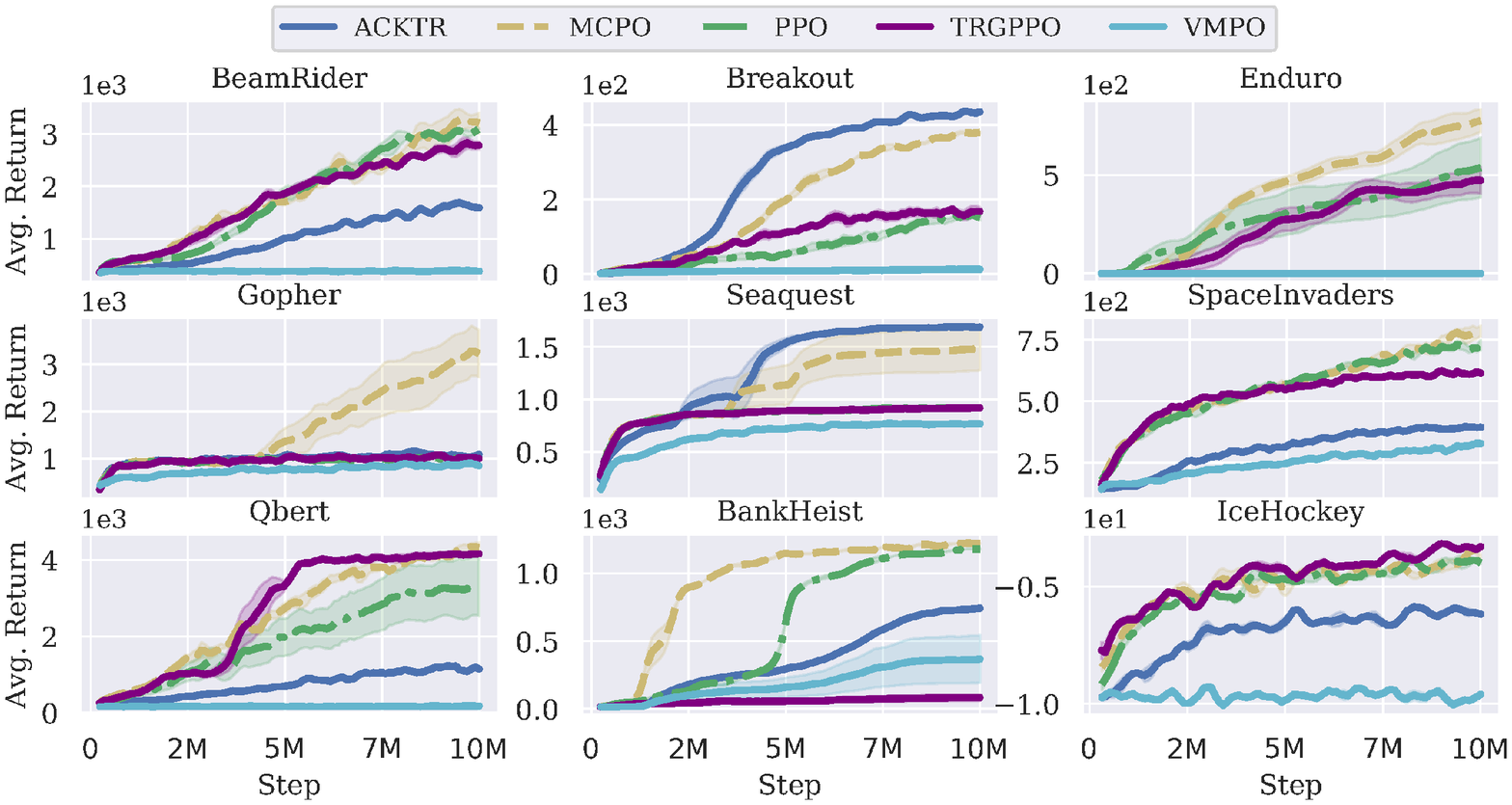}
\par\end{centering}
\caption{Atari games: learning curves (mean and std. over 5 runs) across 10M
training steps.\label{fig:atari10m}}
\end{figure*}

\begin{table}
\begin{centering}
\begin{tabular}{cc}
\hline 
\multirow{1}{*}{Model} & \multicolumn{1}{c}{Chosen hyperparameter}\tabularnewline
\hline 
PPO & $\epsilon=0.2$\tabularnewline
ACKTR & $\delta=0.01$\tabularnewline
VMPO & $\alpha_{0}=5$\tabularnewline
\hline 
MCPO & $N=10$\tabularnewline
\hline 
\end{tabular}
\par\end{centering}
\caption{Signature hyperparameters used in Atari tasks. \label{tab:atarihp-1}}
\end{table}
We also report the average normalized human score (mean and median)
of the models over 6 games in Table \ref{tab:atarinorm-1}. As seen,
MCPO is significantly better than other baselines in terms of both
mean and median normalized human score. We also report full learning
curves of models and normalized human score including TRGPPO in 9
games in Fig. \ref{fig:atari10m} and Table \ref{tab:atarinorm-1},
respectively.

Fig. \ref{fig:atari10m} visualizes the learning curves of the models.
Regardless of our regular tuning, VMPO performs poorly, indicating
that this method is unsuitable or needs extensive tuning to work for
low-sample training regime. ACKTR, works extremely well on certain
games (Breakout and Seaquest), but shows mediocre results on others
(Enduro, BeamRider), overall underperforming MCPO. PPO is always similar
or inferior to MCPO on this testbed. Our MCPO always demonstrates
competitive results, outperforming all other models in 4 games, especially
on Enduro and Gopher, and showing comparable results with that of
the best model in the other 2 games.

\begin{figure*}
\begin{centering}
\includegraphics[width=1\linewidth]{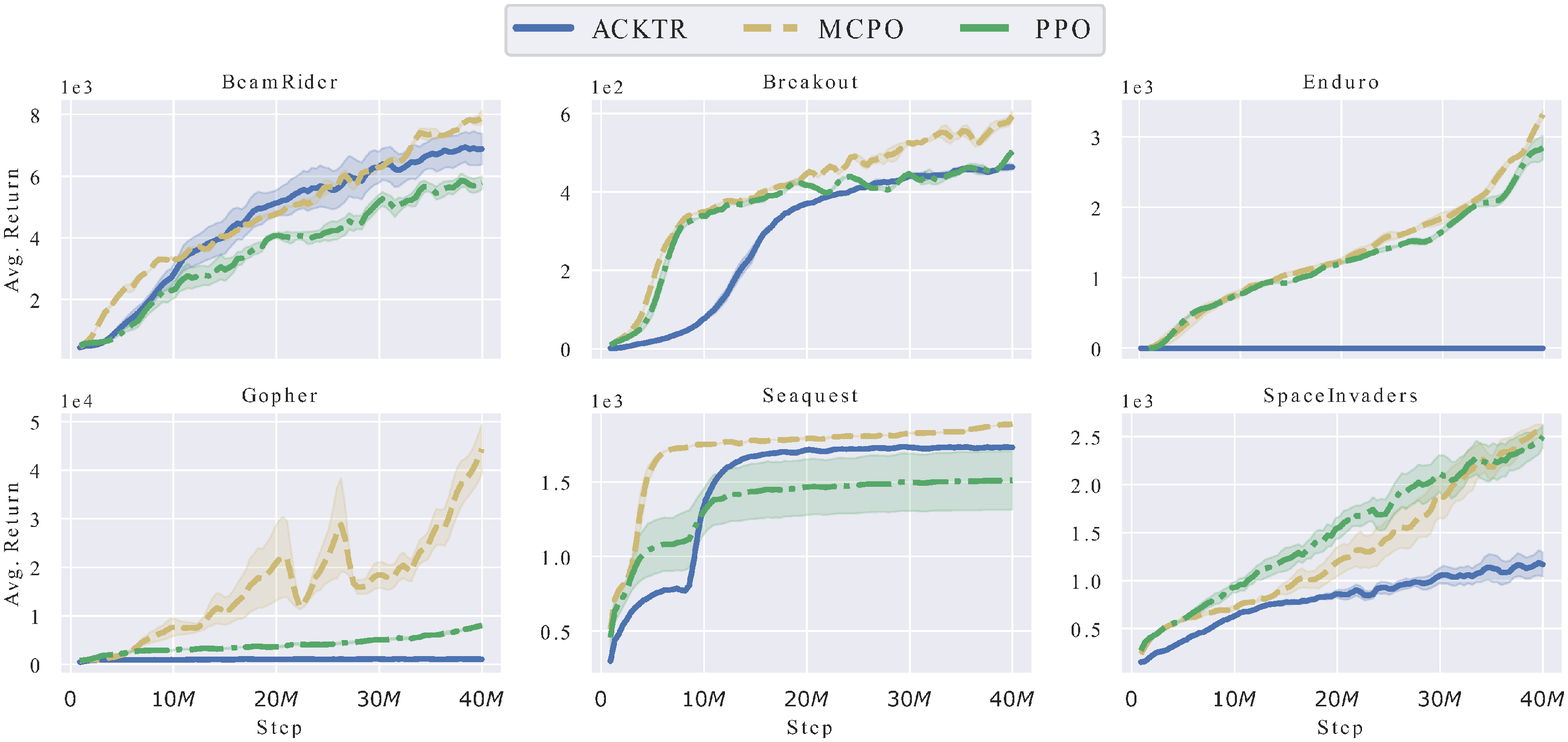}
\par\end{centering}
\caption{Atari games: learning curves (mean and std. over 5 runs) across 40M
training steps.\label{fig:atari40m}}
\end{figure*}

To verify whether MCPO can maintain its performance over longer training,
we examine Atari training for 40 million frames. As shown in Fig.
\ref{fig:atari40m}, MCPO is still the best performer in this training
regime. 

\begin{figure}
\begin{centering}
\includegraphics[width=0.5\columnwidth]{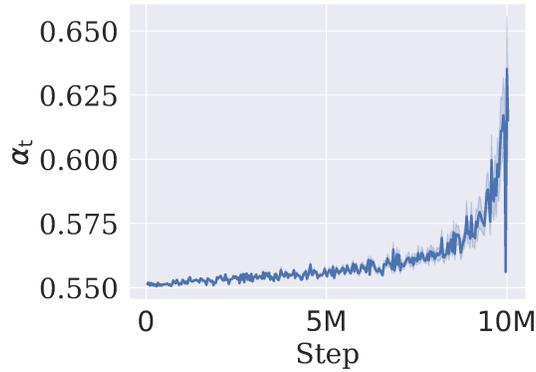}
\par\end{centering}
\caption{HalfCheetah: Average $\alpha_{t}$ over training time (mean and std.
over 3 runs). The training does not use learning rate decay to ensure
that $\psi$ and $\theta_{old}$ do not converge to the same policy
towards the end of training. \label{fig:alphat}}
\end{figure}

\subsubsection{Details on ablation study\label{subsec:Details-on-ablation}}

In this section, we give more details on the ablated baselines. Unless
state otherwise, the baseline use $N=10$.
\begin{itemize}
\item \textbf{No $\psi$} We only changed the objective to

\begin{align}
L_{1}\left(\theta\right) & =\hat{\mathbb{E}}_{t}\left[\tau_{t}\left(\theta\right)\hat{A}_{t}\right]\nonumber \\
 & -\beta\hat{\mathbb{E}}_{t}\left[KL\left[\pi_{\theta_{old}}\left(\cdot|s_{t}\right),\pi_{\theta}\left(\cdot|s_{t}\right)\right]\right]\label{eq:2kl-1}
\end{align}
where $\beta$ is still determined by the $\beta$-switching rule.
This baseline corresponds to setting $\alpha=0$
\item \textbf{Fixed $\alpha=0.5$ }We manually set $\alpha=0.5$ across
training. This baseline uses both old and virtual policy's trust regions
but with fixed balanced coefficient. 
\item \textbf{Fixed $\alpha=1.0$ }We manually set $\alpha=1.0$ across
training. This baseline only uses virtual policy's trust region.
\item \textbf{Annealed} $\beta$ We determine the $\beta$ in Eq. \ref{eq:2kl}
by MDPO's annealing rule, a.k.a, $\beta_{i}=1.0-\frac{i}{T_{total}}$
where $T_{total}$ is the total number of training policy update steps
and $i$ is the current update step. We did not test with other rules
such as fixed or adaptive $\beta$ as we realize that MDPO is often
better than KL Fixed and KL Adaptive in our experiments, indicating
that the annealed $\beta$ is a stronger baseline. 
\item \textbf{Adaptive} $\beta$ We adopt adaptive $\beta$, determined
by the rule introduced in PPO paper (adaptive KL) with $d_{targ}=0.03$.
\item \textbf{Frequent writing }We add a new policy to $\mathcal{M}$ at
every policy update step.
\item \textbf{Uniform writing }Inspired by the uniform writing mechanism
in Memory-Augmented Neural Networks \cite{le2019learning}, we add
a new policy to $\mathcal{M}$ at every interval of 10 update steps.
The interval size could be tuned to get better results but it would
require additional effort, so we preferred our diversity-promoting
writing over this one. 
\item \textbf{Sparse writing }Uniform writing with interval of 100 update
steps.
\item \textbf{Mean $\psi$ }The virtual policy is determined as 
\end{itemize}
\begin{equation}
\psi=\sum_{j}^{\left|\mathcal{M}\right|}\theta_{j}\label{eq:new_psi-1}
\end{equation}

\begin{itemize}
\item \textbf{Half feature }We only use features from 1 to 6 listed in Table
\ref{tab:Features-of-the}.
\end{itemize}
The other baselines including KL Adaptive, KL Fixed, MDPO, PPO, and
VMPO are the same as in \ref{subsec:Details-on-Classical}. The full
learning curves of all models with different hyperparameters are plotted
in Fig. \ref{fig:abl}.


\subsection{Theoretical analysis of MCPO\label{subsec:Theoretical-analysis-of}}

In this section, we explain the design of our objective function $L_{1}$
and $L_{2}$. We want to emphasize that the two trust regions (corresponding
to $\theta_{old}$ and $\psi$) are both important for MCPO's convergence.
Eq. \ref{eq:2kl} needs to include the old policy's trust region because,
in theory, constraining policy updates using the last sampling policy's
trust region guarantees monotonic improvement \cite{schulman2015trust}.
However, in practice, the old policy can be suboptimal and may not
induce much improvement. This motivates us to employ an additional
trust region to regulate the update in case the old policy's trust
region is bad. In doing so, we still want to maintain the theoretical
property of trust-region update while enabling a more robust optimization
that works well even when the ideal setting for theoretical assurance
does not hold. 

It should be noted that constraining with a policy different from
the sampling one would break the monotonic improvement property of
trust-region update. Fortunately, our theory proves that using the
two trust regions as defined in our paper helps maintain the monotonic
improvement property. This is an important result since if you use
an arbitrary virtual policy to form the second trust region (unlike
the one we suggest in this paper), the property may not hold.

Similar to \cite{schulman2015trust}, we can construct a theoretically
guaranteed version of our practical objective functions that ensures
monotonic policy improvement. 

First, we explain the design of $L_{1}$ by recasting $L_{1}$ as

\begin{align*}
L_{1\theta_{old}}\left(\theta\right) & =L_{\theta_{old}}\left(\theta\right)\\
 & -C_{1}D_{KL}^{max}\left(\theta_{old},\theta\right)\\
 & -C_{2}D_{KL}^{max}\left(\psi,\theta\right)
\end{align*}
where $L_{\theta_{old}}\left(\theta\right)=\eta\left(\pi_{\theta_{old}}\right)+\sum_{s}\rho_{\pi_{\theta_{old}}}\left(s\right)\sum_{a}\pi_{\theta}\left(a|s\right)A_{\pi_{\theta_{old}}}\left(s,a\right)$--the
local approximation to the expected discounted return $\eta\left(\theta\right)$,
$D_{KL}^{max}\left(a,b\right)=\max_{s}KL\left[\pi_{a}\left(\cdot|s\right),\pi_{b}\left(\cdot|s\right)\right]$,
$C_{1}=\frac{4\max_{s,a}\left|A_{\pi}\left(s,a\right)\right|\gamma}{\left(1-\gamma\right)^{2}}$
and $C_{2}>0$. Here, $\rho_{\pi_{\theta_{old}}}$ is the (unnormalized)
discounted visitation frequencies induced by the policy $\pi_{\theta_{old}}$.

As the KL is non-negative, $L_{1\theta_{old}}\left(\theta\right)\leq L_{\theta_{old}}\left(\theta\right)-C_{1}D_{KL}^{max}\left(\theta_{old},\theta\right)$.
According to \cite{schulman2015trust}, the RHS is a lower bound on
$\eta\left(\theta\right)$, so $L_{1}$ is also a lower bound on $\eta\left(\theta\right)$
and thus, it is reasonable to maximize the practical $L_{1}$, which
is an approximation of $L_{1\theta_{old}}$.

Next, we show that by optimizing both $L_{1}$ and $L_{2}$, we can
interpret our algorithm as a monotonic policy improvement procedure.
As such, we need to reformulate $L_{2}$ as

\[
L_{2\theta_{old}}\left(\psi\right)=L_{\theta_{old}}\left(\psi\right)-C_{1}D_{KL}^{max}\left(\theta_{old},\psi\right)
\]

Note that compared to the practical $L_{2}$ (as defined in the main
paper on page 5), we have introduced here an additional $KL$ term,
which means we need to find $\psi$ that is close to $\theta_{old}$
and maximizes the approximate advantage $L_{\theta_{old}}\left(\psi\right)$.
As we maximize $L_{2\theta_{old}}\left(\psi\right)$, the maximizer
$\psi$ satisfies 

\[
L_{2\theta_{old}}\left(\psi\right)\geq L_{2\theta_{old}}\left(\theta_{old}\right)=L_{\theta_{old}}\left(\theta_{old}\right)
\]

We also have 

\begin{align}
\eta\left(\theta\right) & \geq L_{1\theta_{old}}\left(\theta\right)\label{eq:l1ie}\\
\eta\left(\theta_{old}\right) & =L_{\theta_{old}}\left(\theta_{old}\right)\leq L_{2\theta_{old}}\left(\psi\right)\nonumber \\
 & =L_{\theta_{old}}\left(\psi\right)-C_{1}D_{KL}^{max}\left(\theta_{old},\psi\right)\nonumber \\
 & =L_{1\theta_{old}}\left(\psi\right)\label{eq:l2ie}
\end{align}

Subtracting both sides of Eq. \ref{eq:l2ie} from Eq. \ref{eq:l1ie}
yields

\[
\eta\left(\theta\right)-\eta\left(\theta_{old}\right)\geq L_{1\theta_{old}}\left(\theta\right)-L_{1\theta_{old}}\left(\psi\right)
\]
Thus by maximizing $L_{1\theta_{old}}\left(\theta\right)$, we guarantee
that the true objective $\eta\left(\theta\right)$ is non-decreasing.

Although the theory suggests that the optimal $L_{2}$ could be $L_{2}^{*}=\hat{\mathbb{E}}_{t}\left[R_{t}\left(\psi_{\varphi}\right)-C_{1}KL\left[\pi_{\theta_{old}}\left(\cdot|s_{t}\right),\pi_{\psi}\left(\cdot|s_{t}\right)\right]\right]$,
it would require additional tuning of $C_{1}$. More importantly,
optimizing an objective in form of $L_{2}^{*}$ needs a very small
step size, and could converge slowly. Hence, we simply discard the
KL term and only optimize $L_{2}=\hat{\mathbb{E}}_{t}\left[R_{t}\left(\psi_{\varphi}\right)\right]$
instead. Empirical results show that using this simplification, MCPO's
learning curves still generally improve monotonically over training
time. 

%% file: main.bbl
\begin{thebibliography}{31}
\providecommand{\natexlab}[1]{#1}
\providecommand{\url}[1]{\texttt{#1}}
\expandafter\ifx\csname urlstyle\endcsname\relax
  \providecommand{\doi}[1]{doi: #1}\else
  \providecommand{\doi}{doi: \begingroup \urlstyle{rm}\Url}\fi

\bibitem[Abdolmaleki et~al.(2018)Abdolmaleki, Springenberg, Tassa, Munos,
  Heess, and Riedmiller]{abdolmaleki2018maximum}
Abbas Abdolmaleki, Jost~Tobias Springenberg, Yuval Tassa, Remi Munos, Nicolas
  Heess, and Martin Riedmiller.
\newblock Maximum a posteriori policy optimisation.
\newblock In \emph{International Conference on Learning Representations}, 2018.

\bibitem[Bertsekas and Tsitsiklis(1995)]{bertsekas1995neuro}
Dimitri~P Bertsekas and John~N Tsitsiklis.
\newblock Neuro-dynamic programming: an overview.
\newblock In \emph{Proceedings of 1995 34th IEEE conference on decision and
  control}, volume~1, pages 560--564. IEEE, 1995.

\bibitem[Blundell et~al.(2016)Blundell, Uria, Pritzel, Li, Ruderman, Leibo,
  Rae, Wierstra, and Hassabis]{blundell2016model}
Charles Blundell, Benigno Uria, Alexander Pritzel, Yazhe Li, Avraham Ruderman,
  Joel~Z Leibo, Jack Rae, Daan Wierstra, and Demis Hassabis.
\newblock Model-free episodic control.
\newblock \emph{arXiv preprint arXiv:1606.04460}, 2016.

\bibitem[Chevalier-Boisvert et~al.(2018)Chevalier-Boisvert, Willems, and
  Pal]{gym_minigrid}
Maxime Chevalier-Boisvert, Lucas Willems, and Suman Pal.
\newblock Minimalistic gridworld environment for openai gym.
\newblock \url{https://github.com/maximecb/gym-minigrid}, 2018.

\bibitem[Fakoor et~al.(2020)Fakoor, Chaudhari, and Smola]{fakoor2020p3o}
Rasool Fakoor, Pratik Chaudhari, and Alexander~J Smola.
\newblock P3o: Policy-on policy-off policy optimization.
\newblock In \emph{Uncertainty in Artificial Intelligence}, pages 1017--1027.
  PMLR, 2020.

\bibitem[Kakade and Langford(2002)]{Kakade2002ApproximatelyOA}
S.~Kakade and J.~Langford.
\newblock Approximately optimal approximate reinforcement learning.
\newblock In \emph{ICML}, 2002.

\bibitem[Lazic et~al.(2021)Lazic, Hao, Abbasi-Yadkori, Schuurmans, and
  Szepesvari]{lazic2021optimization}
Nevena Lazic, Botao Hao, Yasin Abbasi-Yadkori, Dale Schuurmans, and Csaba
  Szepesvari.
\newblock Optimization issues in kl-constrained approximate policy iteration.
\newblock \emph{arXiv preprint arXiv:2102.06234}, 2021.

\bibitem[Le and Venkatesh(2022)]{le2020neurocoder}
Hung Le and Svetha Venkatesh.
\newblock Neurocoder: General-purpose computation using stored neural programs.
\newblock In \emph{Proceedings of the 39th International Conference on Machine
  Learning}, 2022.

\bibitem[Le et~al.(2019{\natexlab{a}})Le, Tran, and Venkatesh]{le2019learning}
Hung Le, Truyen Tran, and Svetha Venkatesh.
\newblock Learning to remember more with less memorization.
\newblock \emph{arXiv preprint arXiv:1901.01347}, 2019{\natexlab{a}}.

\bibitem[Le et~al.(2019{\natexlab{b}})Le, Tran, and Venkatesh]{le2019neural}
Hung Le, Truyen Tran, and Svetha Venkatesh.
\newblock Neural stored-program memory.
\newblock In \emph{International Conference on Learning Representations},
  2019{\natexlab{b}}.

\bibitem[Le et~al.(2021)Le, George, Abdolshah, Tran, and
  Venkatesh]{le2021modelbased}
Hung Le, Thommen~Karimpanal George, Majid Abdolshah, Truyen Tran, and Svetha
  Venkatesh.
\newblock Model-based episodic memory induces dynamic hybrid controls.
\newblock In A.~Beygelzimer, Y.~Dauphin, P.~Liang, and J.~Wortman Vaughan,
  editors, \emph{Advances in Neural Information Processing Systems}, 2021.
\newblock URL \url{https://openreview.net/forum?id=Z9Kpr38Kx_}.

\bibitem[Lillicrap et~al.(2016)Lillicrap, Hunt, Pritzel, Heess, Erez, Tassa,
  Silver, and Wierstra]{lillicrap2016continuous}
Timothy~P Lillicrap, Jonathan~J Hunt, Alexander Pritzel, Nicolas Heess, Tom
  Erez, Yuval Tassa, David Silver, and Daan Wierstra.
\newblock Continuous control with deep reinforcement learning.
\newblock In \emph{ICLR (Poster)}, 2016.

\bibitem[Mnih et~al.(2013)Mnih, Kavukcuoglu, Silver, Graves, Antonoglou,
  Wierstra, and Riedmiller]{mnih2013playing}
Volodymyr Mnih, Koray Kavukcuoglu, David Silver, Alex Graves, Ioannis
  Antonoglou, Daan Wierstra, and Martin Riedmiller.
\newblock Playing atari with deep reinforcement learning.
\newblock \emph{arXiv preprint arXiv:1312.5602}, 2013.

\bibitem[Mnih et~al.(2015)Mnih, Kavukcuoglu, Silver, Rusu, Veness, Bellemare,
  Graves, Riedmiller, Fidjeland, Ostrovski, et~al.]{mnih2015human}
Volodymyr Mnih, Koray Kavukcuoglu, David Silver, Andrei~A Rusu, Joel Veness,
  Marc~G Bellemare, Alex Graves, Martin Riedmiller, Andreas~K Fidjeland, Georg
  Ostrovski, et~al.
\newblock Human-level control through deep reinforcement learning.
\newblock \emph{nature}, 518\penalty0 (7540):\penalty0 529--533, 2015.

\bibitem[Mnih et~al.(2016)Mnih, Badia, Mirza, Graves, Lillicrap, Harley,
  Silver, and Kavukcuoglu]{mnih2016asynchronous}
Volodymyr Mnih, Adria~Puigdomenech Badia, Mehdi Mirza, Alex Graves, Timothy
  Lillicrap, Tim Harley, David Silver, and Koray Kavukcuoglu.
\newblock Asynchronous methods for deep reinforcement learning.
\newblock In \emph{International conference on machine learning}, pages
  1928--1937. PMLR, 2016.

\bibitem[Pirotta et~al.(2013)Pirotta, Restelli, Pecorino, and
  Calandriello]{pirotta2013safe}
Matteo Pirotta, Marcello Restelli, Alessio Pecorino, and Daniele Calandriello.
\newblock Safe policy iteration.
\newblock In \emph{International Conference on Machine Learning}, pages
  307--315. PMLR, 2013.

\bibitem[Schulman et~al.(2015{\natexlab{a}})Schulman, Levine, Abbeel, Jordan,
  and Moritz]{schulman2015trust}
John Schulman, Sergey Levine, Pieter Abbeel, Michael Jordan, and Philipp
  Moritz.
\newblock Trust region policy optimization.
\newblock In \emph{International conference on machine learning}, pages
  1889--1897. PMLR, 2015{\natexlab{a}}.

\bibitem[Schulman et~al.(2015{\natexlab{b}})Schulman, Moritz, Levine, Jordan,
  and Abbeel]{schulman2015high}
John Schulman, Philipp Moritz, Sergey Levine, Michael Jordan, and Pieter
  Abbeel.
\newblock High-dimensional continuous control using generalized advantage
  estimation.
\newblock \emph{arXiv preprint arXiv:1506.02438}, 2015{\natexlab{b}}.

\bibitem[Schulman et~al.(2017)Schulman, Wolski, Dhariwal, Radford, and
  Klimov]{schulman2017proximal}
John Schulman, Filip Wolski, Prafulla Dhariwal, Alec Radford, and Oleg Klimov.
\newblock Proximal policy optimization algorithms.
\newblock \emph{arXiv preprint arXiv:1707.06347}, 2017.

\bibitem[Shani et~al.(2020)Shani, Efroni, and Mannor]{shani2020adaptive}
Lior Shani, Yonathan Efroni, and Shie Mannor.
\newblock Adaptive trust region policy optimization: Global convergence and
  faster rates for regularized mdps.
\newblock In \emph{Proceedings of the AAAI Conference on Artificial
  Intelligence}, volume~34, pages 5668--5675, 2020.

\bibitem[Silver et~al.(2017)Silver, Schrittwieser, Simonyan, Antonoglou, Huang,
  Guez, Hubert, Baker, Lai, Bolton, et~al.]{silver2017mastering}
David Silver, Julian Schrittwieser, Karen Simonyan, Ioannis Antonoglou, Aja
  Huang, Arthur Guez, Thomas Hubert, Lucas Baker, Matthew Lai, Adrian Bolton,
  et~al.
\newblock Mastering the game of go without human knowledge.
\newblock \emph{nature}, 550\penalty0 (7676):\penalty0 354--359, 2017.

\bibitem[Song et~al.(2019)Song, Abdolmaleki, Springenberg, Clark, Soyer, Rae,
  Noury, Ahuja, Liu, Tirumala, et~al.]{song2019v}
H~Francis Song, Abbas Abdolmaleki, Jost~Tobias Springenberg, Aidan Clark,
  Hubert Soyer, Jack~W Rae, Seb Noury, Arun Ahuja, Siqi Liu, Dhruva Tirumala,
  et~al.
\newblock V-mpo: On-policy maximum a posteriori policy optimization for
  discrete and continuous control.
\newblock In \emph{International Conference on Learning Representations}, 2019.

\bibitem[Sutton et~al.(2000)Sutton, McAllester, Singh, and
  Mansour]{sutton2000policy}
Richard~S Sutton, David~A McAllester, Satinder~P Singh, and Yishay Mansour.
\newblock Policy gradient methods for reinforcement learning with function
  approximation.
\newblock In \emph{Advances in neural information processing systems}, pages
  1057--1063, 2000.

\bibitem[Tomar et~al.(2020)Tomar, Shani, Efroni, and
  Ghavamzadeh]{tomar2020mirror}
Manan Tomar, Lior Shani, Yonathan Efroni, and Mohammad Ghavamzadeh.
\newblock Mirror descent policy optimization.
\newblock \emph{arXiv preprint arXiv:2005.09814}, 2020.

\bibitem[Vieillard et~al.(2020{\natexlab{a}})Vieillard, Pietquin, and
  Geist]{vieillard2020deep}
Nino Vieillard, Olivier Pietquin, and Matthieu Geist.
\newblock Deep conservative policy iteration.
\newblock In \emph{Proceedings of the AAAI Conference on Artificial
  Intelligence}, volume~34, pages 6070--6077, 2020{\natexlab{a}}.

\bibitem[Vieillard et~al.(2020{\natexlab{b}})Vieillard, Scherrer, Pietquin, and
  Geist]{vieillard2020momentum}
Nino Vieillard, Bruno Scherrer, Olivier Pietquin, and Matthieu Geist.
\newblock Momentum in reinforcement learning.
\newblock In \emph{International Conference on Artificial Intelligence and
  Statistics}, pages 2529--2538. PMLR, 2020{\natexlab{b}}.

\bibitem[Wang et~al.(2019)Wang, He, Tan, and Gan]{wang2019trust}
Yuhui Wang, Hao He, Xiaoyang Tan, and Yaozhong Gan.
\newblock Trust region-guided proximal policy optimization.
\newblock \emph{Advances in Neural Information Processing Systems},
  32:\penalty0 626--636, 2019.

\bibitem[Wang et~al.(2016)Wang, Bapst, Heess, Mnih, Munos, Kavukcuoglu, and
  de~Freitas]{wang2016sample}
Ziyu Wang, Victor Bapst, Nicolas Heess, Volodymyr Mnih, Remi Munos, Koray
  Kavukcuoglu, and Nando de~Freitas.
\newblock Sample efficient actor-critic with experience replay.
\newblock \emph{arXiv preprint arXiv:1611.01224}, 2016.

\bibitem[Watkins and Dayan(1992)]{watkins1992q}
Christopher~JCH Watkins and Peter Dayan.
\newblock Q-learning.
\newblock \emph{Machine learning}, 8\penalty0 (3-4):\penalty0 279--292, 1992.

\bibitem[Wu et~al.(2017)Wu, Mansimov, Liao, Grosse, and Ba]{wu2017scalable}
Yuhuai Wu, Elman Mansimov, Shun Liao, Roger Grosse, and Jimmy Ba.
\newblock Scalable trust-region method for deep reinforcement learning using
  kronecker-factored approximation.
\newblock In \emph{Proceedings of the 31st International Conference on Neural
  Information Processing Systems}, pages 5285--5294, 2017.

\bibitem[Yang et~al.(2019)Yang, Zheng, Zhang, Zhang, Zheng, Wen, and
  Pan]{yang2019policy}
Long Yang, Gang Zheng, Haotian Zhang, Yu~Zhang, Qian Zheng, Jun Wen, and Gang
  Pan.
\newblock Policy optimization with stochastic mirror descent.
\newblock \emph{arXiv preprint arXiv:1906.10462}, 2019.

\end{thebibliography}
